\documentclass[nojss,shortnames]{jss}

\author{N. Benjamin Erichson \\
		University of Washington 
		\And        
		Ariana Mendible \\
		University of Washington 
        \AND
        Sophie Wihlborn \\
        Fidelity International
        \And
        J. Nathan Kutz \\ 
        University of Washington
   } 
\title{Randomized Nonnegative Matrix Factorization}

\Plainauthor{N. Benjamin Erichson, Ariana Mendible, Sophie Wihlborn, J. Nathan Kutz} 
\Plaintitle{Randomized Nonnegative Matrix Factorization} 
\Shorttitle{Randomized Nonnegative Matrix Factorization} 

\Abstract{
Nonnegative matrix factorization (NMF) is a powerful tool for data mining. However, the emergence of `big data' has severely challenged our ability to compute this fundamental decomposition using deterministic algorithms. This paper presents a randomized hierarchical alternating least squares (HALS) algorithm to compute the NMF. By deriving a smaller matrix from the nonnegative input data, a more efficient nonnegative decomposition can be computed. Our algorithm scales to big data applications while attaining a near-optimal factorization. The proposed algorithm is evaluated using synthetic and real world data and shows substantial speedups compared to deterministic HALS. 
}
\Plainkeywords{dimension reduction, randomized algorithm, nonnegative matrix factorization} 

\Address{
	N. Benjamin Erichson\\
	Department of Applied Mathematics\\
	University of Washington \\
	Seattle, USA\\
	E-mail: \email{erichson@uw.edu} \\
}

\usepackage[utf8]{inputenc} 
\usepackage[T1]{fontenc}    
\usepackage{booktabs}       
\usepackage{amsfonts}       
\usepackage{nicefrac}       
\usepackage{microtype}      

\usepackage{amsmath}
\usepackage{amssymb}
\usepackage{graphicx}
\usepackage{epstopdf}
\usepackage[justification=centering]{caption}
\usepackage{subcaption}
\usepackage{multirow}
\usepackage{float}
\usepackage{setspace}
\usepackage{rotating}
\usepackage{wrapfig,lipsum}
\usepackage{listings}
\usepackage{paralist}
\usepackage{bm}
\usepackage{overpic}

\usepackage{tabularx}
\usepackage{tabulary}

\usepackage[]{algorithm2e}

\usepackage{amsthm}

\theoremstyle{definition}

\theoremstyle{remark}
\newtheorem{remark}{Remark}

\usepackage{tikz}
\usetikzlibrary{matrix,positioning,shapes,shadows,arrows,calc,decorations.pathreplacing}
\usetikzlibrary{arrows.meta}

\tikzstyle{blocky} = [draw, line width=1.2pt,fill=white, rectangle, 
minimum height=2.5em, minimum width=4.5em, rounded corners]    

\tikzstyle{blocky2} = [draw, line width=1.2pt,fill=red!10, rectangle, 
minimum height=2.5em, minimum width=18.5em, rounded corners]

\definecolor{dkgray}{rgb}{99,99,99}
\definecolor{darkred}{RGB}{228,26,28}
\definecolor{darkblue}{RGB}{44,127,184}
\definecolor{magentaCB}{RGB}{221,28,119}

\definecolor{morange}{RGB}{255, 187, 0}
\definecolor{mblue}{RGB}{ 0, 161, 241}

\newcommand{\bA}{\mathbf{A}}

\newcommand{\bQ}{\mathbf{Q}}
\newcommand{\bB}{\mathbf{B}}

\usepackage{kbordermatrix}

\renewcommand{\H}{\mathbf{H}}

\newcommand{\W}{\mathbf{W}}

\newcommand{\X}{\mathbf{X}}

\DeclareMathOperator{\Tr}{Tr}

\begin{document}

\section[Introduction]{Introduction}
Techniques for dimensionality reduction, such as principal component analysis (PCA), are essential to the analysis of high-dimensional data.
These methods take advantage of redundancies in the data in order to find a low-rank and parsimonious model describing the underlying structure of the input data. Indeed, at the core of machine learning is the assumption that low-rank structures are embedded in high-dimensional data~\citep{udell2017nice}.
Dimension reduction techniques find basis vectors which represent the data as a linear combination in lower-dimensional space. This enables the identification of key features and efficient analysis of high-dimensional data. 

A significant drawback of PCA and other commonly-used dimensionality reduction techniques is that they permit both positive and negative terms in their components. 
In many data analysis applications, negative terms fail to hold physically meaningful interpretation.  
For example, images are represented as a grid of nonnegative pixel intensity values. In this context, the negative terms in principal components have no interpretation. 

To address this problem, researchers have proposed restricting the set of basis vectors to nonnegative terms~\citep{paatero1994positive,lee1999learning}.
The paradigm is called nonnegative matrix factorization (NMF) and it has emerged as a powerful dimension reduction technique.
This versatile tool allows computation of sparse (parts-based) and physically meaningful factors that describe coherent structures within the data.
Prominent applications of NMF are in the areas of image processing, information retrieval and gene expression analysis, see for instance the surveys by~\cite{berry2007algorithms} and~\cite{gillis2014and}.
However, NMF is computationally intensive and becomes infeasible for massive data.
Hence, innovations that reduce computational demands are increasingly important in the era of `big data'. 

Randomized methods for linear algebra have been recently introduced to ease the computational demands posed by classical matrix factorizations~\citep{Mahoney2011,RandNLA}. 
\cite{wang2010efficient} proposed to use random projections to efficiently compute the NMF.
Later,~\cite{tepper2016compressed} proposed compressed NMF algorithms based on the idea of bilateral random projections~\citep{zhou2012bilateral}. 
While these compressed algorithms reduce the computational load considerably, they often fail to converge in our experiments. 

We follow the probabilistic approach for matrix approximations formulated by~\cite{halko2011rand}. Specifically, we propose a randomized hierarchical alternating least squares (HALS) algorithm to compute the NMF. 
We demonstrate that the randomized algorithm eases the computational challenges posed by massive data, assuming that the input data feature low-rank structure. Experiments show that our algorithm is reliable and attains a near-optimal factorization.
Further, this manuscript is accompanied by the open-software package \textit{ristretto}, written in \textit{Python}, which allows the reproduction of all results (GIT repository: \url{https://github.com/erichson/ristretto}).  

The manuscript is organized as follows: 
First, Section~\ref{sec:background} briefly reviews the NMF as well as the basic concept of randomized matrix algorithms. 
Then, Section~\ref{sec:rNMF} describes a randomized variant of the HALS algorithm. This is followed by an empirical evaluation in Section~\ref{sec:results}, where synthetic and real world data are used to demonstrate the performance of the algorithm. 
Finally, Section~\ref{sec:conclusion} concludes the manuscript.

\section{Background}\label{sec:background}

\subsection{Low-Rank Matrix Factorization}\label{sec:lowrank}

Low-rank approximations are fundamental and widely used tools for data analysis, dimensionality reduction, and data compression. 
The goal of these methods is to find two matrices of much lower rank that approximate a high-dimensional matrix $\mathbf{X}$:  
\begin{equation}
\begin{array}{cccc}
\mathbf{X} & \approx & \mathbf{W} & \mathbf{H}. \\
m\times n &   &  m\times k & k\times n
\end{array}\label{eq:lowrank} 
\end{equation}
The target rank of the approximation is denoted by $k$, an integer between $1$ and $\text{min}\{m,n\}$.
A ubiquitous example of these tools, the singular value decomposition (SVD), finds the exact solution to this problem in a least-square sense~\citep{Eckart1936psych}. 
While the optimality property of the SVD and similar methods is desirable in many scientific applications, the resulting factors are not guaranteed to be physically meaningful in many others. 
This is because the SVD  imposes orthogonality constraints on the factors, leading to a holistic, rather than parts-based, representation of the input data. 
Further, the basis vectors in the SVD and other popular decompositions are mixed in sign.
 
Thus, it is natural to formulate alternative factorizations which may not be optimal in a least-square sense, but which may preserve useful properties such as sparsity and nonnegativity.
Such properties are found in the NMF.

\subsection{Nonnegative Matrix Factorization}\label{sec:nmf}
The roots of NMF can be traced back to the work by \cite{paatero1994positive}.
\cite{lee1999learning} independently introduced and popularized the concept of NMF in the context of psychology several years later. 

Formally, the NMF attempts to solve Equation~\eqref{eq:lowrank} with the additional nonnegativity constraints: $\mathbf{W}\geq 0$ and $\mathbf{H}\geq 0$. 
These constraints enforce that the input data are expressed as an additive linear combination.
This leads to sparse, parts-based features appearing in the decomposition, which have an intuitive interpretation. For example, NMF components of a face image dataset reveal individual nose and mouth features, whereas PCA components yield holistic features, known as `eigenfaces'. 

Though NMF bears the desirable property of interpretability, the optimization problem is inherently nonconvex and ill-posed. In general, no convexification exists to simplify the optimization, meaning that no exact or unique solution is guaranteed~\citep{gillis2017introduction}. Different NMF algorithms, therefore, can produce distinct decompositions that minimize the objective function. 

We refer the reader to~\cite{lee2001algorithms},~\cite{berry2007algorithms}, and~\cite{gillis2014and} for a comprehensive discussion of the NMF and its applications.
There are two main classes of NMF algorithms~\citep{gillis2017introduction}, discussed in the following.

\subsubsection{Standard Nonlinear Optimization Schemes}

Traditionally, the challenge of finding the nonnegative factors is formulated as the following optimization problem:
\begin{equation}\label{eq:nmf_opt}
\begin{aligned}
& \underset{}{\text{minimize}}
& & f(\mathbf{W}, \mathbf{H}) = \|\mathbf{X-WH}\|_F^2 \\
& \text{subject to}
& & {\bf W\geq 0} \text{ and } {\bf H\geq 0}.
\end{aligned}
\end{equation}
Here $\|\cdot\|_F$ denotes the Frobenius norm of a matrix.
However, the optimization problem in Equation~\eqref{eq:nmf_opt} is nonconvex with respect to both the factors $\mathbf{W}$ and $\mathbf{H}$. 
To resolve this, most NMF algorithms divide the problem into two simpler subproblems which have closed-form solutions.
The convex subproblem is solved by keeping one factor fixed while updating the other, alternating and iterating until convergence. 
One of the most popular techniques to minimize the subproblems is the method of multiplicative updates (MU) proposed by~\cite{lee1999learning}. This procedure is essentially a rescaled version of gradient descent. Its simple implementation comes at the expense of a much slower convergence. 

More appealing are alternating least squares methods and their variants. Among these, HALS proves to be highly efficient~\citep{cichocki2009fast}. 
Without being exhaustive, we would also like to point out the interesting work by~\cite{gillis2012accelerated} as well as by~\cite{kim2014algorithms} who proposed improved and accelerated HALS algorithms for computing NMF. 

\subsubsection{Separable Schemes}

Another approach to compute NMF is based on the idea of column subset selection. In this method, $k$ columns of the input matrix are chosen to form the factor matrix $\mathbf{W} := \mathbf{X}(:,J)$, where $J$ denotes the index set.
The factor matrix $\mathbf{H}$ is found by solving the following optimization problem:
\begin{equation}\
\begin{aligned}
& \underset{}{\text{minimize}}
& & f(\mathbf{H}) = \|\mathbf{X}-\mathbf{W}(:,J)\mathbf{H}\|_F^2 
& \text{s.t.}
& & {\bf H\geq 0}.
\end{aligned}
\end{equation} 
In context of NMF, this approach is appealing if the input matrix is separable~\citep{arora2012computing}. 
This means it must be possible to select basis vectors for $\mathbf{W}$ from the columns of the input matrix $\mathbf{X}$. 
In this case, selecting actual columns from $\mathbf{X}$ preserves the underlying structure of the data and allows a meaningful interpretation.
This assumption is intrinsic in many applications, e.g., document classification and blind hyperspectral unmixing~\citep{gillis2017introduction}. However, this approach has limited potential in applications where the data is dense or noisy. 

These separable schemes are not unique and can be obtained through various algorithms. Finding a meaningful column subset is explored in the CX decomposition~\citep{boutsidis2014near}, which extracts columns that best describe the data. 
In addition, the CUR decomposition~\citep{mahoney2009cur} leverages statistical significance of both columns and rows to improve interpretability, leading to near-optimal decompositions. Another interesting algorithm to compute the near-separable NMF was proposed by~\cite{zhou2013divide}. This algorithm finds conical hulls in which smaller subproblems can be computed in parallel in 1D or 2D. For details on ongoing research in column selection algorithms, we refer the reader  to~\cite{wang2013improving},~\cite{boutsidis2017optimal}, and~\cite{wang2016towards}.

\subsection{Probabilistic Framework}\label{sec:framework}

In the era of `big data', probabilistic methods have become indispensable for computing low-rank matrix approximations. 
The  central concept is to utilize randomness in order to form a surrogate matrix which captures the essential information of a high-dimensional input matrix. This assumes that the input matrix features low-rank structure, i.e., the effective rank is smaller than its ambient dimensions. 
We refer the reader to the surveys by~\cite{halko2011rand},~\cite{Mahoney2011},~\cite{RandNLA} and~\cite{RandomizedMatrixComputations} for more detailed discussions of randomized algorithms. 
For implementations details, for instance, see~\cite{szlam2014implementation},~\cite{voronin2015rsvdpack}, and~\cite{erichson2016randomized}.

Following~\cite{halko2011rand}, the probabilistic framework for low-rank approximations proceeds as follows. 
Let $\mathbf{A}$ be an $m\times n$ matrix, without loss of generality we assume that $n\leq m$.
First, we aim to approximate the range of $\mathbf{A}$. While the SVD provides the best possible basis in a least-square sense, a near-optimal basis can be obtained using random projections
\begin{equation}\label{eq:YXC}
\mathbf{Y} := \mathbf{A}\mathbf{\Omega},
\end{equation}
where $\mathbf{\Omega} \in \mathbb{R}^{n\times k}$ is a random test matrix. Recall, that the target rank of the approximation is denoted by the integer $k$, and is assumed to be $k \ll n$. 
Typically, the entries of $\mathbf{\Omega}$ are independently and identically drawn from the standard normal distribution. Next, the QR-decomposition of $\mathbf{Y}$ is used to form a matrix $\mathbf{Q} \in \mathbb{R}^{m\times k}$ with orthogonal columns. Thus, this matrix forms a near-optimal normal basis for the input matrix such that
\begin{equation}\label{eq:AQQA}
\mathbf{A} \approx \mathbf{Q}\mathbf{Q}^\top\mathbf{A}
\end{equation}
is satisfied. Finally, a smaller matrix $\mathbf{B} \in \mathbb{R}^{k\times n}$ is computed by projecting the input matrix to low-dimensional space  
\begin{equation}
\mathbf{B} := \mathbf{Q}^\top\mathbf{A}.
\end{equation}
Hence, the input matrix can be approximately decomposed (also called QB decomposition) as 
\begin{equation}
 \mathbf{A} \approx \mathbf{Q}\mathbf{B}.
\end{equation}
This process preserves the geometric structure in a Euclidean sense. The smaller matrix $\mathbf{B}$ is, in many applications, sufficient to construct a desired low-rank approximation. 
The approximation quality can be controlled by oversampling and the use of power iterations. 
Oversampling is required to find a good basis matrix. This is because real-world data often do not have exact rank. Thus, instead of just computing $k$ random projections we compute $k+p$ random projections in order to form the basis matrix $\mathbf{Y}$. Specifically, this procedure increases the probability that $\mathbf{Y}$ approximately captures the column space of $\mathbf{X}$. Our experiments show that small oversampling values of about $p=\{10,20\}$ achieve good approximation results to compute the NMF. 
Next, the idea of power iterations is to pre-process the input matrix in order to sample from a matrix which has a faster decaying singular value spectrum~\citep{rokhlin2009randomized}. 
Therefore, Equation~\eqref{eq:YXC} is replaced by
\begin{equation}
\mathbf{Y} = \big( ( \mathbf{X}\mathbf{X}^* )^{q} \mathbf{X} \big)  \mathbf{\Omega}
\end{equation}
where $q$ specifies the number of power iterations. The drawback to this method, is that additional passes over the input matrix are required. Note that the direct implementation of power iteration is numerically unstable, thus, subspace iterations are used instead~\citep{gu2015subspace}.

\subsubsection{Scalability}

We can construct the basis matrix $\mathbf{Q}$ using a deterministic algorithm when the data fit into fast memory. However, deterministic algorithms can become infeasible for data which are too big to fit into fast memory.  
Randomized methods for linear algebra provide a scalable architecture, which ease some of the challenges posed by big data. One of the key advantages of randomized methods is pass efficiency, i.e., the number of complete passes required over the entire data matrix. 

The randomized algorithm, which is sketched above, requires only two passes over the entire data matrix, compared to $k$ passes required by deterministic methods. Hence, the smaller matrix $\mathbf{B}$ can be efficiently constructed if a subset of rows or columns of the data can be accessed efficiently. Specifically, we can construct $\mathbf{Y}$ by sequentially reading in columns or blocks of consecutive columns. See, Appendix~\ref{appendix:qb} for more details and a prototype algorithm. Note, that there exist also single pass algorithms to construct $\mathbf{B}$~\citep{tropp2016randomized}, however, the performance depends substantially on the singular value spectrum of the data. Thus, we favor the slightly more expensive multi-pass framework. Also, see the work by~\cite{elvar} for an interesting discussion on the pass efficiency of randomized methods.

The randomized framework can be also extended to distributed and parallel computing. \cite{voronin2015rsvdpack} proposed a blocked scheme
to compute the QB decomposition in parallel. Using this algorithm, $\mathbf{B}$ can be constructed by distributing the data across processors which have no access to a shared memory to exchange
information. 

\subsubsection{Theoretical Performance of the QB Decomposition}
~\cite{RandomizedMatrixComputations} provides the following simplified description of the expected error: 
\begin{equation*}
\mathop{\mathbb{E}}\| \bA - \bQ\bB  \|_2  \leq  \Bigg[ 1 + \sqrt{\frac{k}{p-1}} + \frac{e\sqrt{k+p}}{p} \cdot \sqrt{n-k}\Bigg]^\frac{1}{2q+1} \sigma_{k+1}(\bA).
\end{equation*}
It follows that as $p$ increases, the error tends towards the best possible approximation error, i.e., the singular value $\sigma_{k+1}(\bA)$. A rigorous error analysis is provided by \cite{halko2011rand}.

\newpage
\section{Randomized Nonnegative Matrix Factorization}\label{sec:rNMF}

High-dimensional data pose a computational challenge for deterministic nonnegative matrix factorization, despite modern optimization techniques. 
Indeed, the costs of solving the optimization problem formulated in Equation~\eqref{eq:nmf_opt} can be prohibitive.
Our motivation is to use randomness as a strategy to ease the computational demands of extracting low-rank features from high-dimensional data. 
Specifically, a randomized hierarchical alternating least squares (HALS) algorithm is formulated to efficiently compute the nonnegative matrix factorization.

\subsection{Hierarchal Alternating Least Squares}

\begin{figure}[!b]
	\centering	
	\begin{subfigure}[t]{0.45\textwidth}
		\centering
		\DeclareGraphicsExtensions{.pdf}
		\includegraphics[width=1\textwidth]{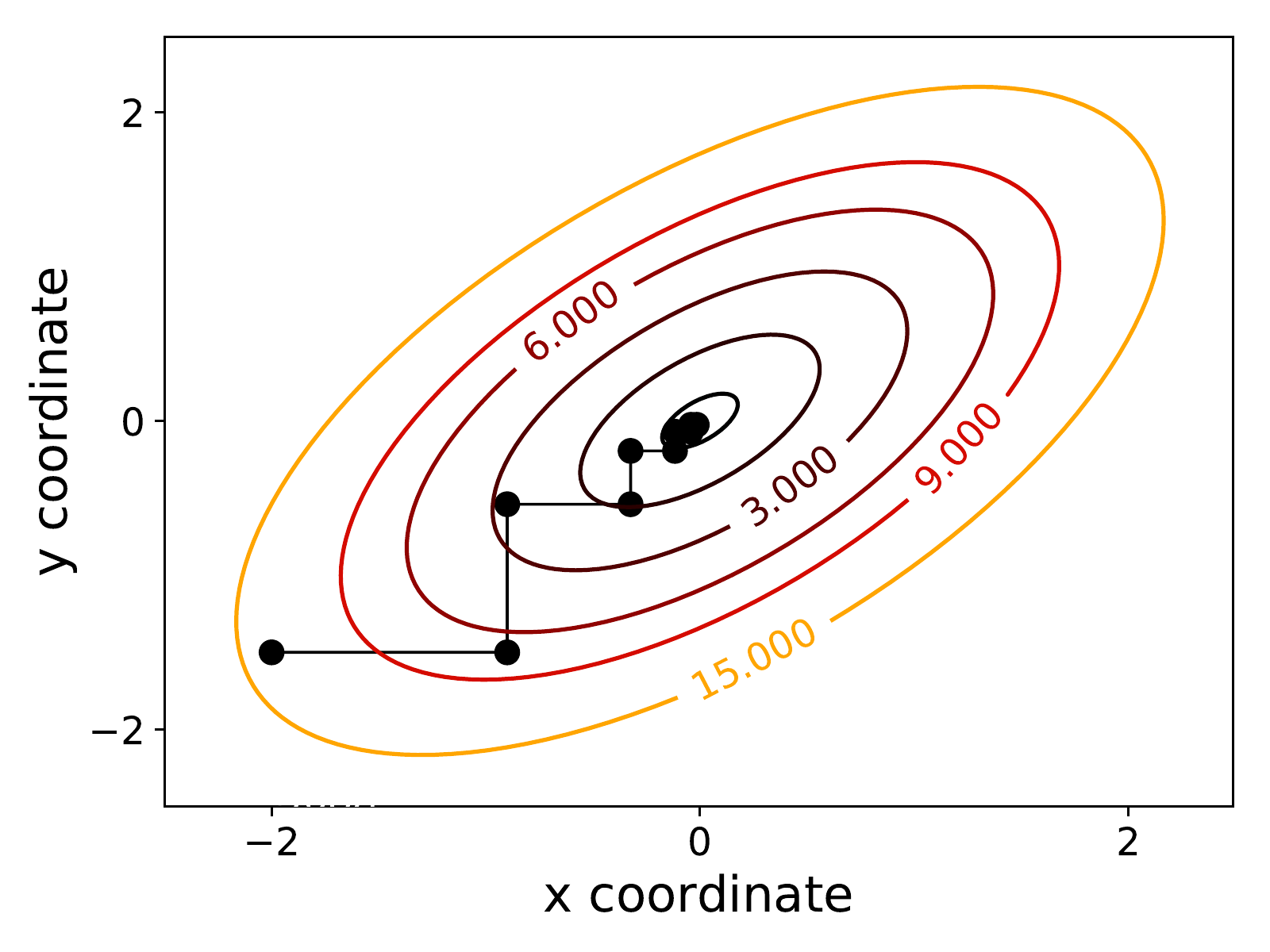}
		\caption{Starting point $x=-2$ and $y=-1.5$. }
	\end{subfigure}	
	\begin{subfigure}[t]{0.45\textwidth}
		\centering
		\DeclareGraphicsExtensions{.pdf}
		\includegraphics[width=1\textwidth]{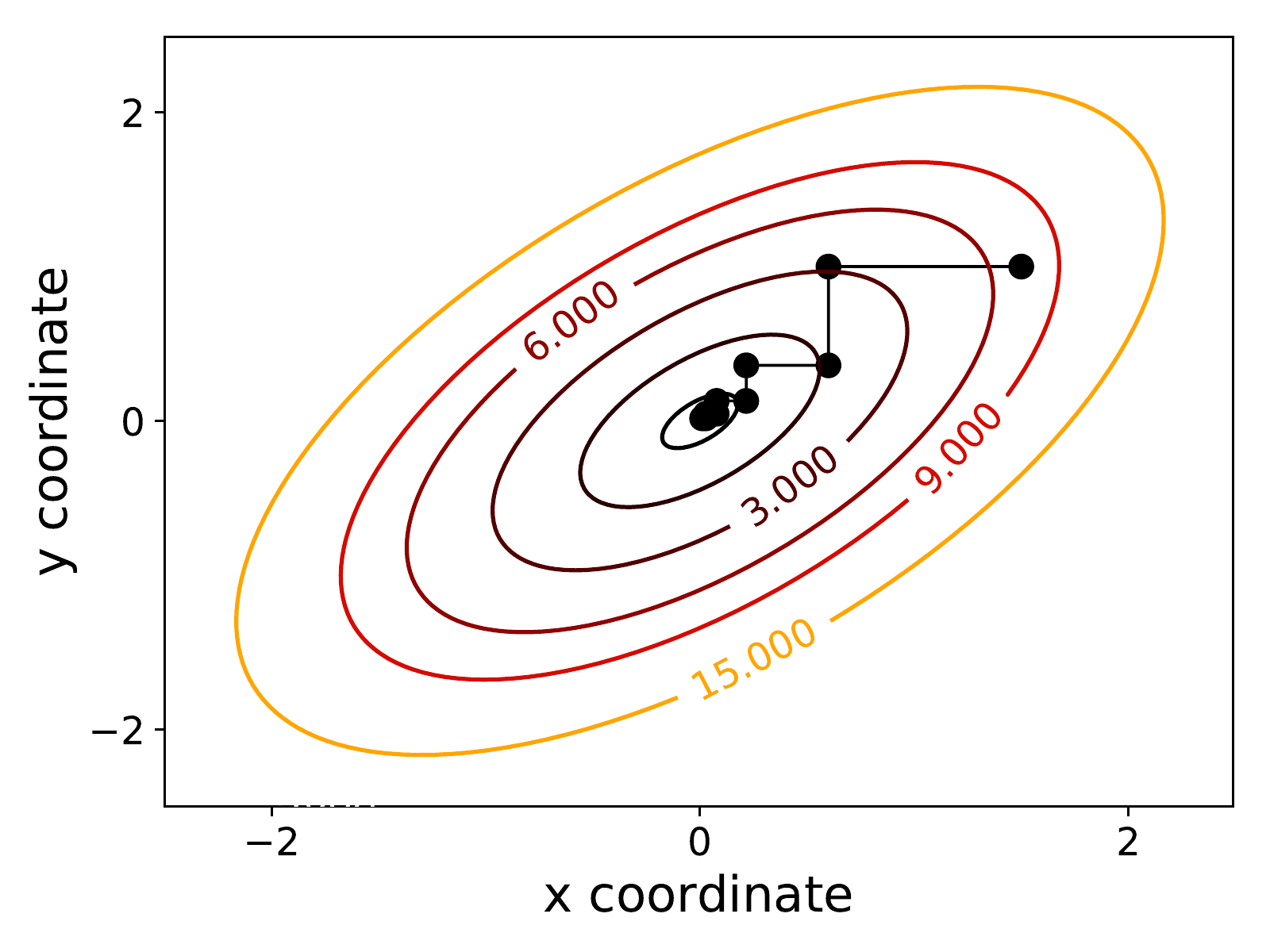}
		\caption{Starting point $x=1.5$ and $y=1.0$. }
	\end{subfigure}	

	\caption{Coordinate descent finds the minimum of a function by successively updating only along one coordinate direction ($x$ or $y$) at a time, while fixing the other direction. In other words, a single variable optimization problem is solved in each iteration. }
	\label{fig:cd_example}
\end{figure}

Block coordinate descent (BCD) methods are a universal approach to algorithmic optimization~\citep{wright2015coordinate}. These iterative methods fix a block of components and optimize with respect to the remaining components.
Figure~\ref{fig:cd_example} illustrates the process for a simple 2-dimensional function $f(x,y)$, where we iteratively update $x$ while $y$ is fixed
\begin{equation*}
x = x + \nu \dfrac{\partial f}{\partial x}(x,y),
\end{equation*}
and $y$ while $x$ is fixed
\begin{equation*}
y = y + \nu \dfrac{\partial f}{\partial y}(x,y),
\end{equation*}
until convergence is reached. The parameter $\nu$ controls the step size and can be chosen in various ways~\citep{wright2015coordinate}.  
Following this philosophy, the HALS algorithm unbundles the original problem into a sequence of simpler optimization problems. 
This allows the efficient computation of the NMF~\citep{cichocki2009fast}. 

Suppose that we update $\mathbf{W}$ and $\mathbf{H}$ by fixing most terms except for the block comprised of the $j$th column $\mathbf{W}_{(:,j)}$ and $j$th row $\mathbf{H}_{(j,:)}$. 
Thus, each subproblem is essentially reduced to a smaller minimization.
HALS approximately minimizes the cost function in Equation~\eqref{eq:nmf_opt} with respect to the remaining $k-1$ components
\begin{equation}\label{eq:hals}
\begin{aligned}
& \underset{}{\text{minimize}}
& & J_j({\bf W}_{(:,j)},  
	{\bf H}_{(j,:)}) = \|\mathbf{R}^{(j)}-\mathbf{W}_{(:,j)}\mathbf{H}_{(j,:)}\|_F^2,
\end{aligned}
\end{equation}
where $\mathbf{R}^{(j)}$ is the $j$th residual
\begin{equation}  \label{eq:residual} 
\mathbf{R}^{(j)} := X - \sum_{i \neq j}^{k} {\bf W}_{(:,i)}{\bf H}_{(i,:)}.
\end{equation}
This can be viewed as a decomposition of the residual~\citep{pmlr-v39-kimura14}.
Then, it is simple to derive the gradients to find the stationary points for both components.
First, we expend the cost function in Eq.~\eqref{eq:hals} as
\begin{align*}
J_j
&= \Tr \left [ \mathbf{R}^{(j)\top} \mathbf{R}^{(j)} - 2 \mathbf{R}^{(j)\top}  \W_{(:,j)} \H_{(j,:)} +  \H_{(j,:)} ^\top \W_{(:,j)}^\top    \W_{(:,j)} \H_{(j,:)}  \right ] \\
&= \|\mathbf{R}^{(j)} \|_F^2 -  2 \Tr \left [ \mathbf{R}^{(j)\top} \W_{(:,j)} \H_{(j,:)} \right ] + \Tr  \left [ \H_{(j,:)} ^\top \W_{(:,j)}^\top    \W_{(:,j)} \H_{(j,:)} \right ]
\end{align*}
Then, we take the gradient of $\W(:,j)$ with respect to $J_j$ 
\begin{equation*}  
0 = \frac{\partial J_j}{\partial {\bf W}_{(:,j)}} =
{\bf W}_{(:,j)}{\bf H}_{(j,:)}{\bf H}^\top_{(j,:)} - \mathbf{R}^{(j)}{\bf H}^\top_{(j,:)},
\end{equation*}
and the gradient of $\H(j,:)$ with respect to $J_j$ 
\begin{equation*} 
0 = \frac{\partial J_j}{\partial {\bf H}_{(j,:)}} =
{\bf H}^\top_{(j,:)}{\bf W}^\top_{(:,j)}{\bf W}_{(:,j)} - \mathbf{R}^{(j)\top}{\bf W}_{(:,j)}.
\end{equation*}
The update rules for the $j$th component of ${\bf W}$ and ${\bf H}$ are
\begin{equation}\label{eq:ruleW}  
{\bf W}_{(:,j)}^+ \gets
\left[\frac{[\mathbf{R}^{(j)}{\bf H}^\top_{(j,:)}}{{\bf H}_{(j,:)}{\bf H}^\top_{(j,:)}} \right]_{+} ,
\end{equation}
\begin{equation}\label{eq:ruleH}   
{\bf H}_{(j,:)}^+ \gets
\left[\frac{\mathbf{R}^{(j)\top}{\bf W}_{(:,j)}}{{\bf W}^\top_{(:,j)}{\bf W}_{(:,j)}} \right]_{+},
\end{equation}
where the maximum operator, defined as $[x]_{+} := \text{max}(0,x)$, ensures that the components remain nonzero. 
Note, that we can express Eq.~\eqref{eq:residual} also as
\begin{align}\label{eq:simpR}
\mathbf{R}^{(j)} &:= \X -  \sum_{i \neq j}^{k} \W_{(:,i)} \H_{(i,:)} = \X -  \W \H +  \W_{(:,j)} \H_{(j,:)}.
\end{align}
Then, Eq.~\ref{eq:simpR} can be substituted into the above update rules in order to avoid the explicit computation of the residual $\mathbf{R}^{(j)}$. Hence, we obtain the following simplified update rules:
\begin{equation}
\mathbf{W}_{(:,j)}^+ \gets
\left[\mathbf{W}_{(:,j)} + \frac{\left[ \mathbf{X}\mathbf{H}^\top\right]_{(:,j)} -  \mathbf{{W}} \left[\mathbf{H}\mathbf{H}^\top\right]_{(:,j)}}{\left[ \mathbf{H}\mathbf{H}^\top\right]_{(j,j)}} \right]_+,
\end{equation}
\begin{equation}
\mathbf{H}_{(j,:)}^+ \leftarrow \left[\mathbf{H}_{(j,:)} + \frac{\left[ \mathbf{X}^\top\mathbf{W}\right]_{(:,j)} - \mathbf{H}^\top \left[\mathbf{W}^\top\mathbf{W} \right]_{(:,j)}}{[\mathbf{W}^\top\mathbf{W}]_{(j,j)}} \right]_+.
\end{equation}

\subsection{Randomized Hierarchal Alternating Least Squares}

Employing randomness, we reformulate the optimization problem in Equation~\eqref{eq:nmf_opt} as a low-dimensional optimization problem. Specifically, the high-dimensional $m \times n$ input matrix $\mathbf{X}$ is replaced by the $k \times n$ surrogate matrix $\mathbf{B}$, which is formed as described in Section~\ref{sec:framework}. Thus we yield the following optimization problem:
\begin{equation}\label{eq:nmf_opt_B}
\begin{aligned}
& \underset{}{\text{minimize}}
& & \tilde{f}(\mathbf{\tilde{W}}, \mathbf{H}) = \|\mathbf{B-\tilde{W}H}\|_F^2 \\
& \text{subject to}
& & {\bf Q\tilde{W}\geq 0} \text{ and } {\bf H\geq 0}.
\end{aligned}
\end{equation}
The nonnegativity constraints need to apply to the high-dimensional factor matrix ${\bf W}$, but not necessarily to ${\bf \tilde{W}}$. 
The matrix ${\bf \tilde{W}}$ can be rotated back to high-dimensional space using the following approximate relationship ${\bf W\approx Q\tilde{W}}$.
Equation~\eqref{eq:nmf_opt_B} can only be solved approximately, since  ${\bf QQ^\top \neq I}$. 
Further, there is no reason that ${\bf \tilde{W}}$ has nonnegative entries, yet the low-dimensional projection will decrease the objective function in Equation~\eqref{eq:nmf_opt_B}.\footnote{A proof can be demonstrated similar to the proof by~\cite{cohen2015fast} for the compressed nonnegative CP tensor decomposition.}
Now, we formulate the randomized HALS algorithm as 
\begin{equation}\label{eq:rhals}
\begin{aligned}
& \underset{}{\text{minimize}}
& & \tilde{J}_j({\bf \tilde{W}}_{(:,j)}, {\bf H}_{(j,:)}) =
\|\mathbf{\tilde{R}}^{(j)}-\mathbf{\tilde{W}}_{(:,j)}\mathbf{H}_{(j,:)}\|_F^2,
\end{aligned}
\end{equation}
where $\mathbf{\tilde{R}}^{(j)}$ is the $j$th compressed residual
\begin{equation}  \label{eq:residualLD} 
\mathbf{\tilde{R}}^{(j)} := \mathbf{B} - \sum_{i \neq j}^{k} {\bf \tilde{W}}_{(:,i)}{\bf H}_{(i,:)}.
\end{equation}
Then, the update rule for the $j$th component of $\mathbf{H}$ is as follows
\begin{equation}
\mathbf{H}_{(j,:)}^+ \leftarrow \left[\mathbf{H}_{(j,:)} + \frac{\left[ \mathbf{B}^\top\mathbf{\tilde{W}}\right]_{(:,j)} - \mathbf{H}^\top \left[ \mathbf{\tilde{W}}^\top\mathbf{\tilde{W}} \right]_{(:,j)}}{[\mathbf{\tilde{W}}^\top\mathbf{\tilde{W}}]_{(j,j)}} \right]_+,
\end{equation}
Note, that we use $[\mathbf{{W}}^\top\mathbf{{W}}]_{(j,j)}$ for scaling in practice in order to ensure the correct scaling in high-dimensional space. Next, the update rule for the  $j$th component of $\mathbf{\tilde{W}}$ is
\begin{equation}
\mathbf{\tilde{W}}_{(:,j)} \gets
\mathbf{\tilde{W}}_{(:,j)} + \frac{\left[ \mathbf{B}\mathbf{H}^\top\right]_{(:,j)} -  \mathbf{\tilde{W}}   \left[\mathbf{H}\mathbf{H}^\top\right]_{(:,j)}}{\left[ \mathbf{H}\mathbf{H}^\top\right]_{(j,j)} }.
\end{equation}
Then, we employ the following scheme to update the $j$th component in high-dimensional space, followed by projecting the updated factor matrix $\mathbf{W}^+$ back to low-dimensional space
\begin{equation}
	{\bf {W}}_{(:,j)}^+ \gets  \lbrack {\bf Q \tilde{W}}_{(:,j)} \rbrack_{+}.		
\end{equation}
\begin{equation}
	{\bf \tilde{W}}_{(:,j)}^+ \gets {\bf Q}^\top {\bf {W}}_{(:,j)}^+ .		
\end{equation}

\begin{figure}[!h]
	\centering	
	\begin{subfigure}[t]{0.85\textwidth}
		\centering
		\DeclareGraphicsExtensions{.pdf}
		\includegraphics[width=1\textwidth]{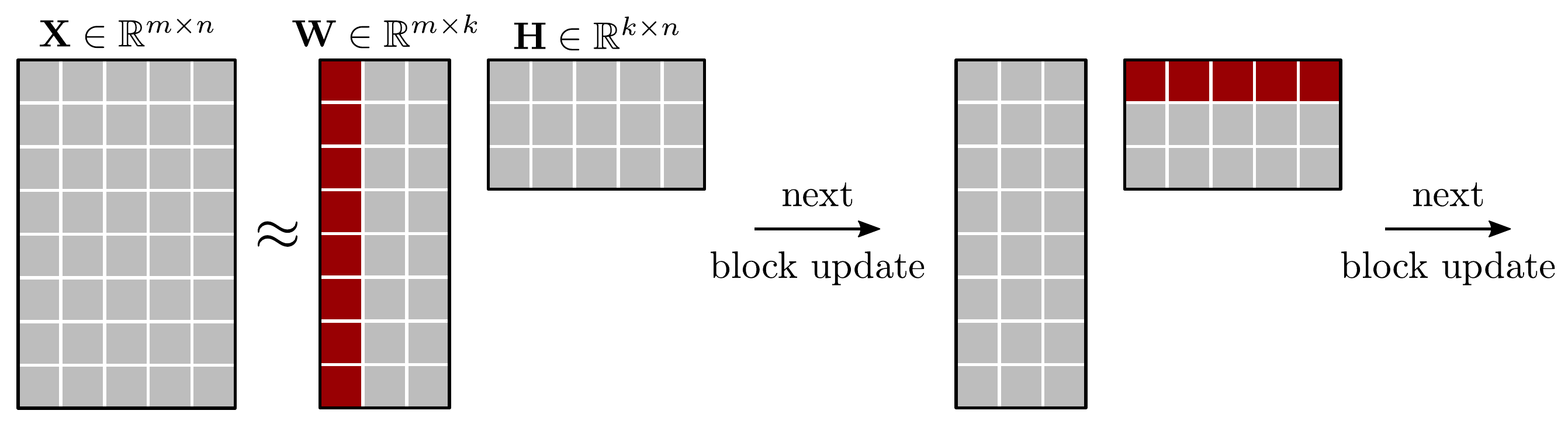}
		\caption{ Update order: ${\bf {W}}_{(:,1)}^+ \rightarrow {\bf {H}}_{(1,:)}^+ \rightarrow  \dots \rightarrow {\bf {W}}_{(:,k)}^+ \rightarrow {\bf {H}}_{(k,:)}^+ $.}
	\end{subfigure}	

	\begin{subfigure}[t]{0.85\textwidth}
		\vspace*{+0.2cm}
		\centering
		\DeclareGraphicsExtensions{.pdf}
		\includegraphics[width=1\textwidth]{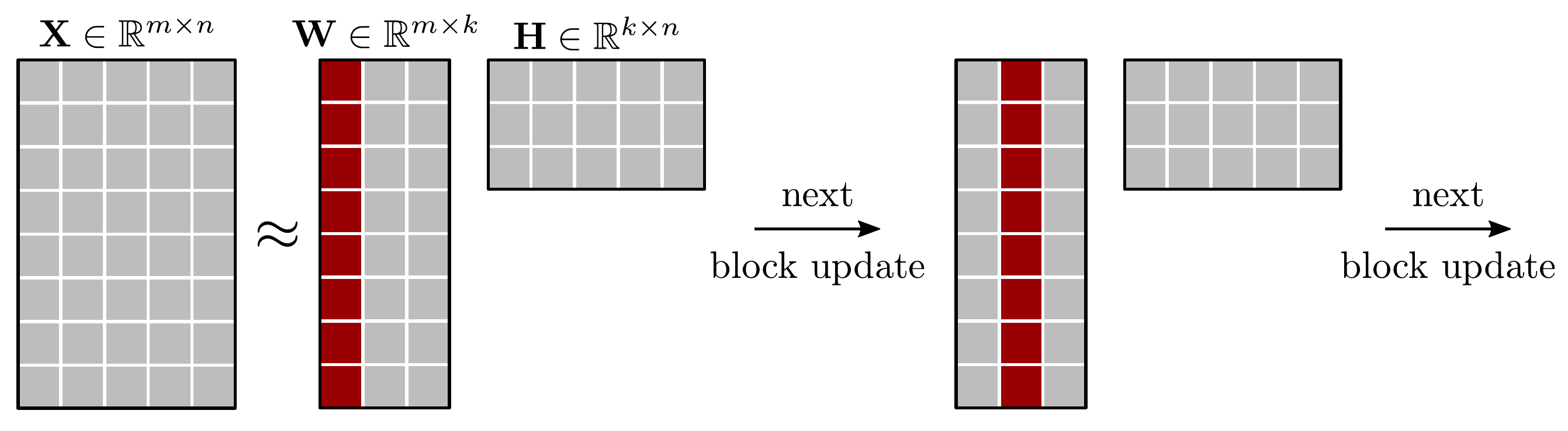}
		\caption{Update order:${\bf {W}}_{(:,1)}^+ \rightarrow {\bf {W}}_{(:,2)}^+ \rightarrow \dots \rightarrow {\bf {W}}_{(:,k)}^+ \rightarrow {\bf {H}}_{(1,:)}^+ \rightarrow {\bf {H}}_{(2,:)}^+ \dots \rightarrow {\bf {H}}_{(k,:)}^+$.}
	\end{subfigure}	
	\caption{Cyclic update orders.}
	\label{fig:update_order}
\end{figure}

This HALS framework allows the choice of different update orders~\citep{kim2014algorithms}. For instance, we can proceed by using one of the following two cyclic update orders:
\begin{equation}
	{\bf {W}}_{(:,1)}^+ \rightarrow {\bf {H}}_{(1,:)}^+ \rightarrow  \dots \rightarrow {\bf {W}}_{(:,k)}^+ \rightarrow {\bf {H}}_{(k,:)}^+,		
\end{equation}
or
\begin{equation}
	{\bf {W}}_{(:,1)}^+ \rightarrow {\bf {W}}_{(:,2)}^+ \rightarrow \dots \rightarrow {\bf {W}}_{(:,k)}^+ \rightarrow {\bf {H}}_{(1,:)}^+ \rightarrow {\bf {H}}_{(2,:)}^+ \dots \rightarrow {\bf {H}}_{(k,:)}^+,		
\end{equation}
which are illustrated in Figure~\ref{fig:update_order}. We favor the latter scheme in the following. Alternatively, a shuffled update order can be used, wherein the components are updated in a random order in each iteration. \cite{wright2015coordinate} noted that this scheme performs better in some applications.  Yet another exciting strategy is randomized block coordinate decent~\citep{nesterov2012efficiency}. 

The computational steps are summarized in Algorithm~\ref{alg:rHALS}.

\subsection{Stopping Criterion}
Predefining a maximum number of iterations is not satisfactory in practice. Instead, we aim to stop the algorithm when a suitable measure has reached some convergence tolerance. For a discussion on stopping criteria, for instance, see~\cite{gillis2012accelerated}. One option is to terminate the algorithm if the value of objective function is smaller than some stopping condition $\epsilon$
\begin{equation}
||\mathbf{X}-\mathbf{W}\mathbf{H}||_F^2 < \epsilon.
\end{equation}   
This criteria is expensive to compute and often not a good measure for convergence. An interesting alternative stopping condition is the projected gradient~\citep{lin2007projected,hsieh2011fast}. The projected gradient $\nabla^P f(\mathbf{W,H})$ measures how close the current solution is to a stationary point. We compute the projected gradient with respect to $\mathbf{W}$ as
\begin{equation}
\nabla^P_{\mathbf{W}_{ij}} :=
\begin{cases}
\dfrac{\partial f(\mathbf{W,H})}{\partial \mathbf{W}_{ij} }, & \text{if}\ \mathbf{W}_{ij} > 0 \\

min(0, \dfrac{\partial f(\mathbf{W,H})}{\partial \mathbf{W}_{ij} }), & \text{if}\ \mathbf{W}_{ij} = 0 \\
\end{cases},
\end{equation}   
and in a similar way with respect to $\mathbf{H}$. Accordingly, the stopping condition can be formulated:
\begin{equation}
||\nabla^P f(\mathbf{W,H})||_F^2 < \epsilon ||\nabla^P f(\mathbf{W^0,H^0})||_F^2,
\end{equation}   
where $\mathbf{W^0}$ and $\mathbf{H^0}$ indicate the initial points. Note, that it follows from the Karush–Kuhn–Tucker (KKT) condition that a stationary point is reached if and only if the projected gradient is  $\nabla^P f(\mathbf{W^*,H^*})=0$ for some optimal points $\mathbf{W^*}$ and $\mathbf{H^*}$.

\begin{algorithm}[!h]
\vspace*{+0.3cm}
	\begin{center}
	\scalebox{0.8}{\fbox{	
		\begin{minipage}{210mm}
			\begin{tabbing}
				\hspace{1mm} \= \hspace{5mm} \= \hspace{2mm} \= \hspace{6mm} \= \hspace{85mm} \=\kill
				\textbf{Require:} A nonnegative matrix $\mathbf{X}$ of dimension $m\times n$, and target rank $k$.\\[1mm]
				\textbf{Optional:} Parameters $p$ and $q$ for oversampling, and power iterations.\\[1mm]

				(1)  \> \> $l = k + p$ \> \> \> {\color{blue}$\textrm{Slight oversampling}$} \\[1mm]
				
				(2)  \> \> $\mathbf{\Omega} = \texttt{rand}(n,l)$ \> \> \> {\color{blue}$\textrm{Generate sampling matrix } \mathbf{\Omega} \in \mathbb{R}^{n\times l}$}\\[1mm]
				
				(3)  \> \> $\mathbf{Y} = \mathbf{X} \mathbf{\Omega}$ \> \> \> {\color{blue}$\textrm{Form basis }\mathbf{Y} \in \mathbb{R}^{m\times l}$}\\[1mm]
				
				(4)  \> \> \textbf{for} $j = 1,\dots,q$ \> \> \> {\color{blue}$\textrm{Optional: subspace iterations}$} \\[1mm]
				
				(5)  \> \> \> $\left[\mathbf{Q},\sim\right] = \texttt{qr}(\mathbf{Y})$ \\[1mm]

				(6)  \> \> \> $\left[\mathbf{Q},\sim\right] = \texttt{qr}(\mathbf{\mathbf{X}^\top \mathbf{Q}})$ \\[1mm]
				
				(7)  \> \> \> $\mathbf{Y} = \mathbf{X} \mathbf{Q}$ \\[2mm]

				(8)  \> \> $\left[\mathbf{Q},\sim\right] = \texttt{qr}(\mathbf{Y})$ \> \> \> {\color{blue}\textrm{Form orthonormal basis } $\mathbf{Q} \in \mathbb{R}^{m\times l}$}
				\\[1mm]
				
				(9)  \> \> $\mathbf{B} = \mathbf{Q}^\top \mathbf{X}$ \> \> \> {\color{blue}$\textrm{Form smaller matrix } \mathbf{B} \in \mathbb{R}^{l\times n}$} \\[1mm]
				
				(10)  \> \> Initialize nonnegative factors $\mathbf{W}\in \mathbb{R}^{m\times k}$,  $\mathbf{\tilde{W}}\in \mathbb{R}^{l\times k}$ and $\mathbf{H}\in \mathbb{R}^{k\times n}$  \> \> \>  \\[1mm]

				(11)  \> \>  \textbf{repeat}  \> \> \>  \\[1mm]
				
				(12)  \> \> \> $\mathbf{R} = \mathbf{B}^\top \mathbf{\tilde{W}}$ \>  \> {\color{blue}$ \mathbf{R} \in \mathbb{R}^{n\times k}$} \\[1mm]  				
				
				(13)  \> \> \> $\mathbf{S} = \mathbf{{W}}^\top \mathbf{{W}}$ \>  \> {\color{blue}$ \mathbf{S} \in \mathbb{R}^{k\times k}$} \\[1mm]

				(14)  \> \> \> \textbf{for} $j = 1,\dots,k$ \> \> {\color{blue} Update $\mathbf{H}$ row by row} \\[1mm]

				(15)  \> \> \> \> $\mathbf{H}(j,:) =  \mathbf{H}(j,:) + (\mathbf{R}(:,j) - \mathbf{H}^\top\mathbf{S}(:,j))/\mathbf{S}(j,j)$ \\[2mm]								
				
				(16)  \> \> \> \> $\mathbf{H}(j,:) = \textbf{max}(0,  \mathbf{H}(j,:) )$ \> {\color{blue} Elementwise maximum operator}\\[2mm]
				
				
				(17)  \> \> \> $\mathbf{T} = \mathbf{B} \mathbf{H}^\top$ \>  \> {\color{blue}$ \mathbf{T} \in \mathbb{R}^{k\times k}$} \\[1mm]  				
				
				(18)  \> \> \> $\mathbf{V} = \mathbf{H} \mathbf{H}^\top$ \>  \> {\color{blue}$ \mathbf{V} \in \mathbb{R}^{k\times k}$} \\[1mm]

				(19)  \> \> \> \textbf{for} $j = 1,\dots,k$ \> \> {\color{blue} Update $\mathbf{W}$ column by column} \\[1mm]

				(20)  \> \> \> \> $\mathbf{\tilde{W}}(:,j) =  \mathbf{\tilde{W}}(:,j)  + (\mathbf{T}(:,j)-\mathbf{\tilde{W}}\mathbf{V}(:,j))/\mathbf{V}(j,j) $ \\[1mm]
				
				(21)  \> \> \> \> $\mathbf{W}(:,j) = \textbf{max} (0,\mathbf{Q}\mathbf{W}(:,j)) $ \> {\color{blue} Elementwise maximum operator} \\[1mm]													
				
				(22)  \> \> \> \> ${\bf \tilde{W}}(:,j) = {\bf Q^\top W}(:,j)$ \> {\color{blue} Rotate to low-dimensional space}\\[2mm]					
				
				(23)  \> \> \> \textbf{if} {stopping criterion or maximum number of iterations is reached}\\[2mm]
				
				\textbf{Return:} Nonnegative factor matrices $\mathbf{W}\in \mathbb{R}^{m\times k}$ and $\mathbf{H}\in \mathbb{R}^{k\times n}$  
				
			\end{tabbing}
		\end{minipage}}}
\end{center}
		\caption{Prototype algorithm to compute the NMF using randomized HALS.}
		\label{alg:rHALS}
		
	\end{algorithm}

	\begin{remark}[Random Test Matrix]
		The entries $\mathbf{\omega}_{ij}$ of the random test matrix $\mathbf{\Omega}$ are drawn independently from the uniform distribution in the interval of $[0,1]$. Nonnegative random entries perform better than Gaussian distributed entries, and seem to be the natural choice in the context of nonnegative data. 
	\end{remark}

	\begin{remark}[Initialization]
		The performance of NMF algorithms depends on the procedure used for initializing the factor matrices. We refer to~\cite{langville2014algorithms} for an excellent discussion on this topic. A standard approach is to initialize the factor matrices with Gaussian entries, where negative elements are set to 0. However, in many applications the performance can be improved by using an initialization scheme which is based on the (randomized) singular value decomposition~\citep{boutsidis2008svd}.	
	\end{remark}

\subsection{Regularized Hierarchal Alternating Least Squares}

\begin{figure}[!hb]
\centering
\begin{subfigure}[t]{0.32\textwidth}
\begin{center}\scalebox{0.7}{
			\begin{tikzpicture}[auto,node distance = 2cm,>=latex']
			
			\coordinate (X1) at (-2,0);
            \coordinate (X2) at (2,0);
            \coordinate (Y1) at (0,-2);
			\coordinate (Y2) at (0,2);
			
			\draw [-latex, black, line width=1.0pt]  (X1) -- (X2);
			\draw [-latex, black, line width=1.0pt]  (Y1) -- (Y2);
			
			\draw[darkred,thick,dashed] (0,0) circle (0.91cm);

			\draw[darkred,thick,] (0,0) circle (0.5cm);

			\draw [-, darkblue, line width=1.0pt]  (-1.5,1.8) -- (1.5,0.3);
           	\draw [fill=black] (0.5,0.8) circle (3pt);


\end{tikzpicture}}
\end{center}
\caption{$\ell_2$ norm. }\label{fig:l2norm}
\end{subfigure}   
\begin{subfigure}[t]{0.32\textwidth}
\begin{center}\scalebox{0.7}{
			\begin{tikzpicture}[auto,node distance = 2cm,>=latex']
			
			\coordinate (X1) at (-2,0);
            \coordinate (X2) at (2,0);
            \coordinate (Y1) at (0,-2);
			\coordinate (Y2) at (0,2);
			
			\draw [-latex, black, line width=1.0pt]  (X1) -- (X2);
			\draw [-latex, black, line width=1.0pt]  (Y1) -- (Y2);
			
			\draw [-, darkred, line width=1.0pt, dashed]  (1,0) -- (0,1);
			\draw [-, darkred, line width=1.0pt, dashed]  (0,1) -- (-1,0);
			\draw [-, darkred, line width=1.0pt, dashed]  (-1,0) -- (0,-1);
			\draw [-, darkred, line width=1.0pt, dashed]  (0,-1) -- (1,0);
			            
			\draw [-, darkred, line width=1.0pt]  (0.5,0) -- (0,0.5);
			\draw [-, darkred, line width=1.0pt]  (0,0.5) -- (-0.5,0);
			\draw [-, darkred, line width=1.0pt]  (-0.5,0) -- (0,-0.5);
			\draw [-, darkred, line width=1.0pt]  (0,-0.5) -- (0.5,0);
			            
			\draw [-, darkblue, line width=1.0pt]  (-1.5,1.8) -- (1.5,0.3);
           	\draw [fill=black] (0,1.02) circle (3pt);


			\end{tikzpicture}}
	\end{center}
\caption{$\ell_1$ norm. }\label{fig:l1norm}
\end{subfigure}    
\begin{subfigure}[t]{0.32\textwidth}
\begin{center}\scalebox{0.7}{
			\begin{tikzpicture}[auto,node distance = 2cm,>=latex']
			
			\coordinate (X1) at (-2,0);
            \coordinate (X2) at (2,0);
            \coordinate (Y1) at (0,-2);
			\coordinate (Y2) at (0,2);
			
			\draw [-latex, black, line width=1.0pt]  (X1) -- (X2);
			\draw [-latex, black, line width=1.0pt]  (Y1) -- (Y2);
			
			\draw [-, darkred, line width=1.0pt, dashed, bend right]  (1,0) edge (0,1);
			\draw [-, darkred, line width=1.0pt, dashed, bend right]  (0,1) edge (-1,0);
			\draw [-, darkred, line width=1.0pt, dashed, bend right]  (-1,0) edge (0,-1);
			\draw [-, darkred, line width=1.0pt, dashed, bend right]  (0,-1) edge (1,0);
                   
			\draw [-, darkred, line width=1.0pt, bend right]  (0.5,0) edge (0,0.5);
			\draw [-, darkred, line width=1.0pt, bend right]  (0,0.5) edge (-0.5,0);
			\draw [-, darkred, line width=1.0pt, bend right]  (-0.5,0) edge (0,-0.5);
			\draw [-, darkred, line width=1.0pt, bend right]  (0,-0.5) edge (0.5,0);
			            
			\draw [-, darkblue, line width=1.0pt]  (-1.5,1.8) -- (1.5,0.3);
           	\draw [fill=black] (0,1.02) circle (3pt);


			\end{tikzpicture}}
	\end{center}
\caption{Elastic net. }\label{fig:enorm}
\end{subfigure}    

\caption{Illustration of popular norms which can be used for regularization, adapted from~\cite{erichson2018sparse}. Both the $\ell_1$ norm and the elastic net have a sparsity-promoting effect. }
\label{Fig:geo_regularizers}. 
\end{figure}
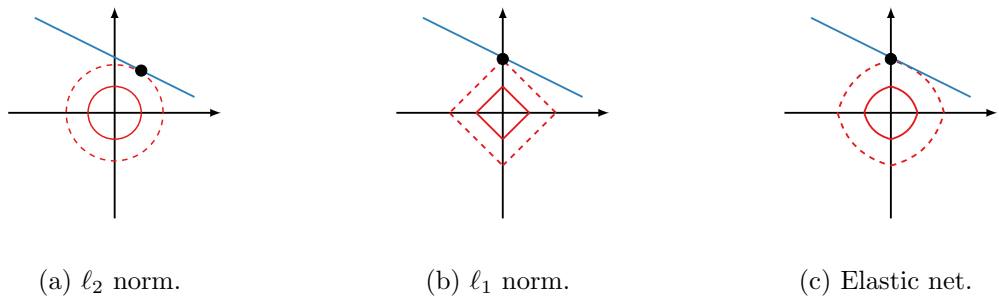

Nonnegative matrix factorization can be extend by incorporating extra constraints such as regularization terms. Specifically, we can formulate the problem more generally as 
\begin{equation}
\begin{aligned}
& \underset{}{\text{minimize}}
& & f(\mathbf{W}, \mathbf{H}) = \|\mathbf{X-WH}\|_F^2 + r_W(\mathbf{W}) + r_H(\mathbf{H}) \\
& \text{subject to}
& & {\bf W\geq 0} \text{ and } {\bf H\geq 0},
\end{aligned}
\end{equation}
where  $r(\cdot)$ is a regularization term. Popular choices for regularization are the $\ell_2$ norm, the $\ell_1$ norm, and a combination of both. The different norms are illustrated in Figure~\ref{Fig:geo_regularizers}. 
 
The $\ell_2$ norm (see Fig.~\ref{fig:l2norm}), or Frobenious norm, as a regularizer adds a small bias against large terms into the updating rules, which is also known as ridge. This regularizer can be defined as
\begin{equation*}
{r}(\bm x) = \alpha \|\bm x \|_{F}^2,
\end{equation*}
where $\alpha$ is a tuning parameter. An additional benefit is less overfitting. 

In many application it is favorable to obtain a more parsimonious model, i.e., a model that is simpler and more interpretable. This can be achieved by using sparsity promoting regularizers which help to identify the most meaningful `active' (non-zero) entries, while forcing many terms to zero. The most popular choice is to use the $\ell_1$ norm as sparsity-promoting regularizer
\begin{equation*}
{r}(\bm x) = \beta \|\bm x \|_{1},
\end{equation*}
which is also known as LASSO (least absolute shrinkage and selection operator), see Fig.~\ref{fig:l1norm}. The tuning parameter $\beta$ can be used to control the level of sparsity. 

A third regularization method, elastic net~\citep{zou2005regularization}, maintains the properties of both ridge and LASSO defined as
\begin{equation*}
{r}(\bm x) = \alpha\|\bm x \|_{1} + \beta\|\bm x \|_{2}^2.
\end{equation*}
Essentially, the elastic net is just a combination of the $\ell_1$ and the Frobenious norm, see Fig.~\ref{fig:enorm}.

The discussed regularizes are simple to integrate into the HALS update rules. For details see~\cite{cichocki2009fast} and \cite{kim2014algorithms}.

\paragraph{Hierarchal Alternating Least Squares with $\ell_2$ Regularization:}

We start by augmenting the cost function with additional $\ell_2$ regularization terms for the components
\begin{equation}
\begin{aligned}
& \underset{}{\text{minimize}}
& & {J}_j({\bf {W}}_{(:,j)},  
{\bf H}_{(j,:)}) = \|\mathbf{{R}}^{(j)}-\mathbf{{W}}_{(:,j)}\mathbf{H}_{(j,:)}\|_F^2 + \alpha_W \|\mathbf{{W}}_{(:,j)} \|_2^2 + \alpha_H \|\mathbf{H}_{(j,:)} \|_2^2,
\end{aligned}
\end{equation}
where $\alpha_W$ and $\alpha_H$ are tuning parameters.
We take the gradient of ${\W}(:,j)$ with respect to ${J}_j$ 
\begin{equation*}  
0 = \frac{\partial {J}_j}{\partial {\bf {W}}_{(:,j)}} =
{\bf {W}}_{(:,j)}{\bf H}_{(j,:)}{\bf H}^\top_{(j,:)} - \mathbf{{R}}^{(j)}{\bf H}^\top_{(j,:)} + \alpha_W {\bf {W}}_{(:,j)},
\end{equation*}
Then, rearranging terms and substituting Eq.~\eqref{eq:residual} yields the following regularized update rule for the $j$th component of ${\bf {W}}$
\begin{equation}
\mathbf{{W}}_{(:,j)}^+ \gets \left[
\dfrac{\left[ \mathbf{H}\mathbf{H}^\top\right]_{(j,j)}}{\left[ \mathbf{H}\mathbf{H}^\top\right]_{(j,j)} + \alpha_W}   \mathbf{{W}}_{(:,j)} + \frac{\left[ \mathbf{X}\mathbf{H}^\top\right]_{(:,j)} -  \mathbf{ {W}} \left[\mathbf{H}\mathbf{H}^\top\right]_{(:,j)}}{\left[ \mathbf{H}\mathbf{H}^\top\right]_{(j,j)} + \alpha_W } \right]_{+}.
\end{equation}
Similar, we yield the following regularized update rule for the components of $\mathbf{H}$
\begin{equation}
\mathbf{H}_{(j,:)}^+ \leftarrow \left[ \dfrac{[\mathbf{{W}}^\top\mathbf{{W}}]_{(j,j)}}{[\mathbf{{W}}^\top\mathbf{\tilde{W}}]_{(j,j)} + \alpha_H }    \mathbf{H}_{(j,:)} + \frac{\left[ \mathbf{X}^\top\mathbf{{W}}\right]_{(:,j)} - \mathbf{H}^\top \left[ \mathbf{{W}}^\top\mathbf{{W}} \right]_{(:,j)}}{[\mathbf{{W}}^\top\mathbf{{W}}]_{(j,j)} + \alpha_H } \right]_+,
\end{equation}

\paragraph{Hierarchal Alternating Least Squares with $\ell_1$ Regularization:}

We start by augmenting the cost function with additional $\ell_1$ regularization terms for the components
\begin{equation}
\begin{aligned}
& \underset{}{\text{minimize}}
& & \tilde{J}_j({\bf \tilde{W}}_{(:,j)},  
{\bf H}_{(j,:)}) = \|\mathbf{\tilde{R}}^{(j)}-\mathbf{\tilde{W}}_{(:,j)}\mathbf{H}_{(j,:)}\|_F^2 + \beta_W \|\mathbf{{W}}_{(:,j)} \|_1 + \beta_H \|\mathbf{H}_{(j,:)} \|_1,
\end{aligned}
\end{equation}
where $\beta_W$ and $\beta_H$ are tuning parameters to control the level of sparsity. Noting, that the entries of $\mathbf{W}_{(:,j)}$ are nonnegative we have that  $\|\mathbf{{W}}_{(:,j)} \|_1 = \sum_{i} \mathbf{W}_{(i,j)}$. Hence, taking the gradient of ${\W}(:,j)$ with respect to ${J}_j$ gives
\begin{equation*}  
0 = \frac{\partial {J}_j}{\partial {\bf {W}}_{(:,j)}} =
{\bf {W}}_{(:,j)}{\bf H}_{(j,:)}{\bf H}^\top_{(j,:)} - \mathbf{{R}}^{(j)}{\bf H}^\top_{(j,:)} + \beta_W \mathbf{1}.
\end{equation*}
Then, we can formulate the following update rule for the $j$th component of ${\W}$ 
\begin{equation}
\mathbf{W}_{(:,j)}^+ \gets
\left[\mathbf{W}_{(:,j)} + \frac{\left[ \mathbf{X}\mathbf{H}^\top - \beta_W \mathbf{1}\right]_{(:,j)} -  \mathbf{{W}} \left[\mathbf{H}\mathbf{H}^\top\right]_{(:,j)}}{\left[ \mathbf{H}\mathbf{H}^\top\right]_{(j,j)}} \right]_+,
\end{equation}
and similar we yield the update rules for the components of $\mathbf{H}$  
\begin{equation}
\mathbf{H}_{(j,:)}^+ \leftarrow \left[\mathbf{H}_{(j,:)} + \frac{\left[ \mathbf{X}^\top\mathbf{W} -  \beta_H \mathbf{1}\right]_{(:,j)} - \mathbf{H}^\top \left[\mathbf{W}^\top\mathbf{W} \right]_{(:,j)}}{[\mathbf{W}^\top\mathbf{W}]_{(j,j)}} \right]_+.
\end{equation}

Both, $\ell_1$ and $\ell_2$ regularization can be combined, which leads to the elastic net~\citep{zou2005regularization}. The elastic net, which combines both the effects of ridge and lasso, shows often a favorable performance in high-dimensional data settings.

\section{Experimental Evaluation}\label{sec:results}

In the following, we evaluate the proposed randomized algorithm and compare it to the deterministic HALS algorithm as implemented in the \textit{scikit-learn} package \citep{scikit-learn}. Througout all experiments, we set the oversampling parameter to $p=20$ and the number of subspace iterations to  $q=2$ for the randomized algorithm. 
%
For comparison, we also provide the results of the compressed MU algorithm~\cite{tepper2016compressed}.\footnote{Note, that~\cite{tepper2016compressed} have also used the active set and the alternating direction method of multipliers for computing the NMF. While both of these methods perform better than the MU algorithm in several empirical experiments, we faced some numerical issues in applications of interest to us. Hence, we present only the results of the compressed MU algorithm in the following.} 
All computations are performed on a laptop with Intel Core i7-7700K CPU Quad-Core 4.20GHz, 64GB fast memory.

\subsection{Facial Feature Extraction}

Feature extraction from facial images is an essential preprocessing step for facial recognition. An ordinary approach is to use the SVD or PCA for this task, which was first studied by \cite{kirby1990application} and later by~\cite{turk1991face}. The resulting basis images represent `shadows' of the faces, the so-called eigenfaces. However, instead of learning holistic representations, it seems more natural to learn a parts-based representation of the facial data. This was first demonstrated in the seminal work by~\cite{lee1999learning} using the NMF. The corresponding basis images are more robust to occlusions and are easier to interpret.  

\begin{figure}[!h]
	\centering
	\begin{subfigure}[t]{0.32\textwidth}
		\centering
		\DeclareGraphicsExtensions{.pdf}
		\includegraphics[width=1\textwidth]{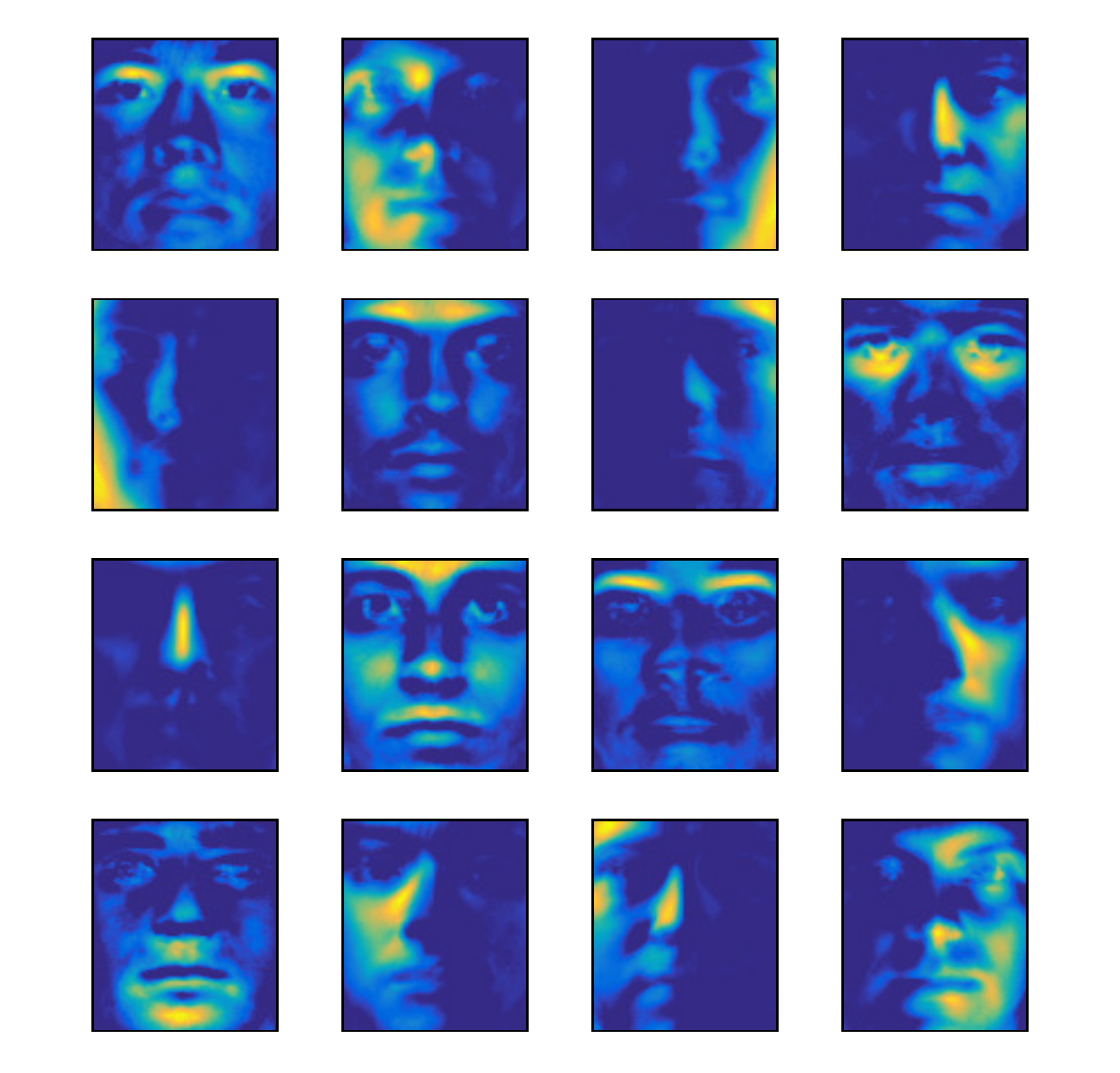}
		\caption{Deterministic NMF. }
	\end{subfigure}
	\begin{subfigure}[t]{0.32\textwidth}
		\centering
		\DeclareGraphicsExtensions{.pdf}
		\includegraphics[width=1\textwidth]{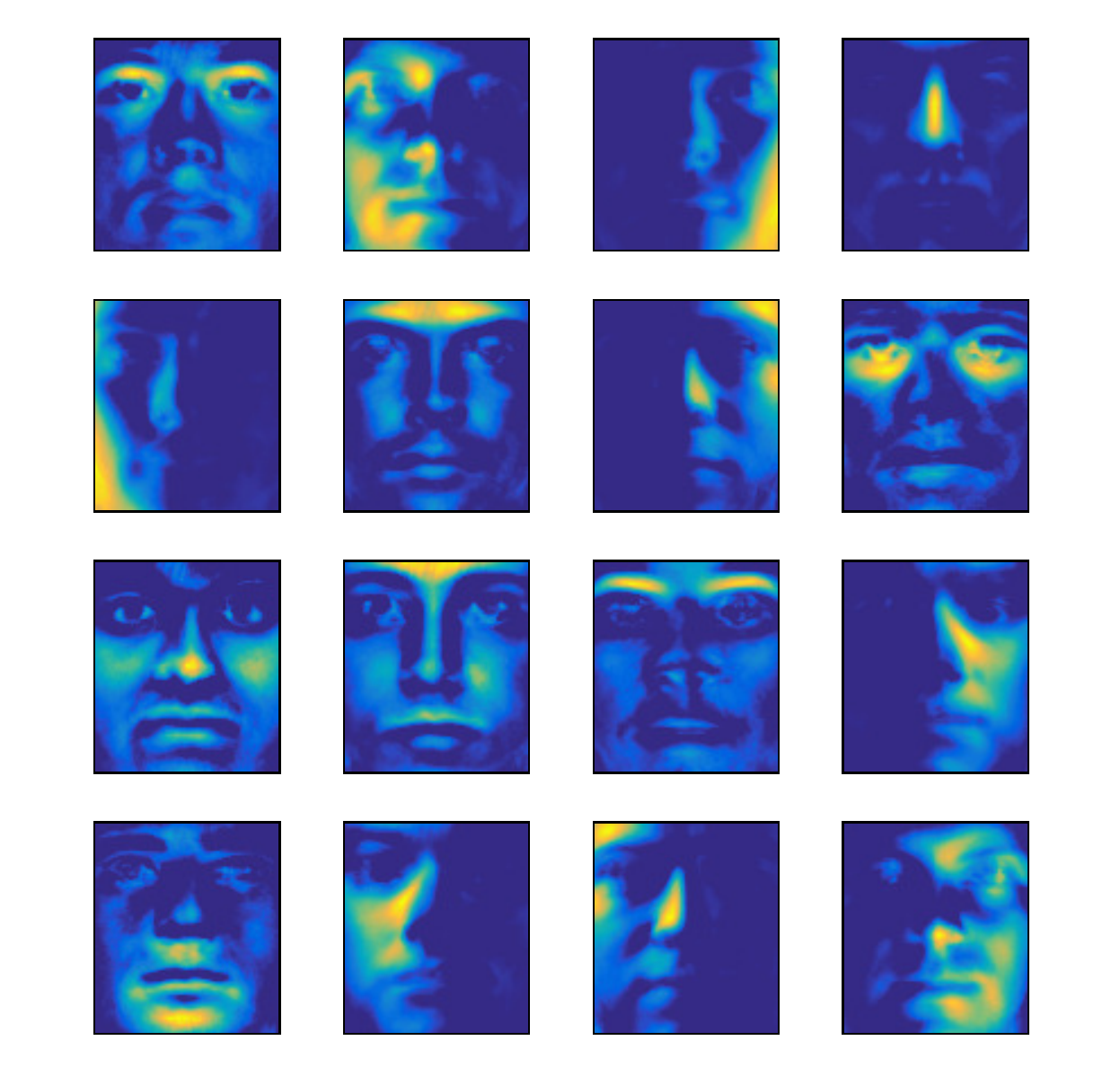}
		\caption{Randomized NMF.}
	\end{subfigure}
	\begin{subfigure}[t]{0.32\textwidth}
		\centering
		\DeclareGraphicsExtensions{.pdf}
		\includegraphics[width=1\textwidth]{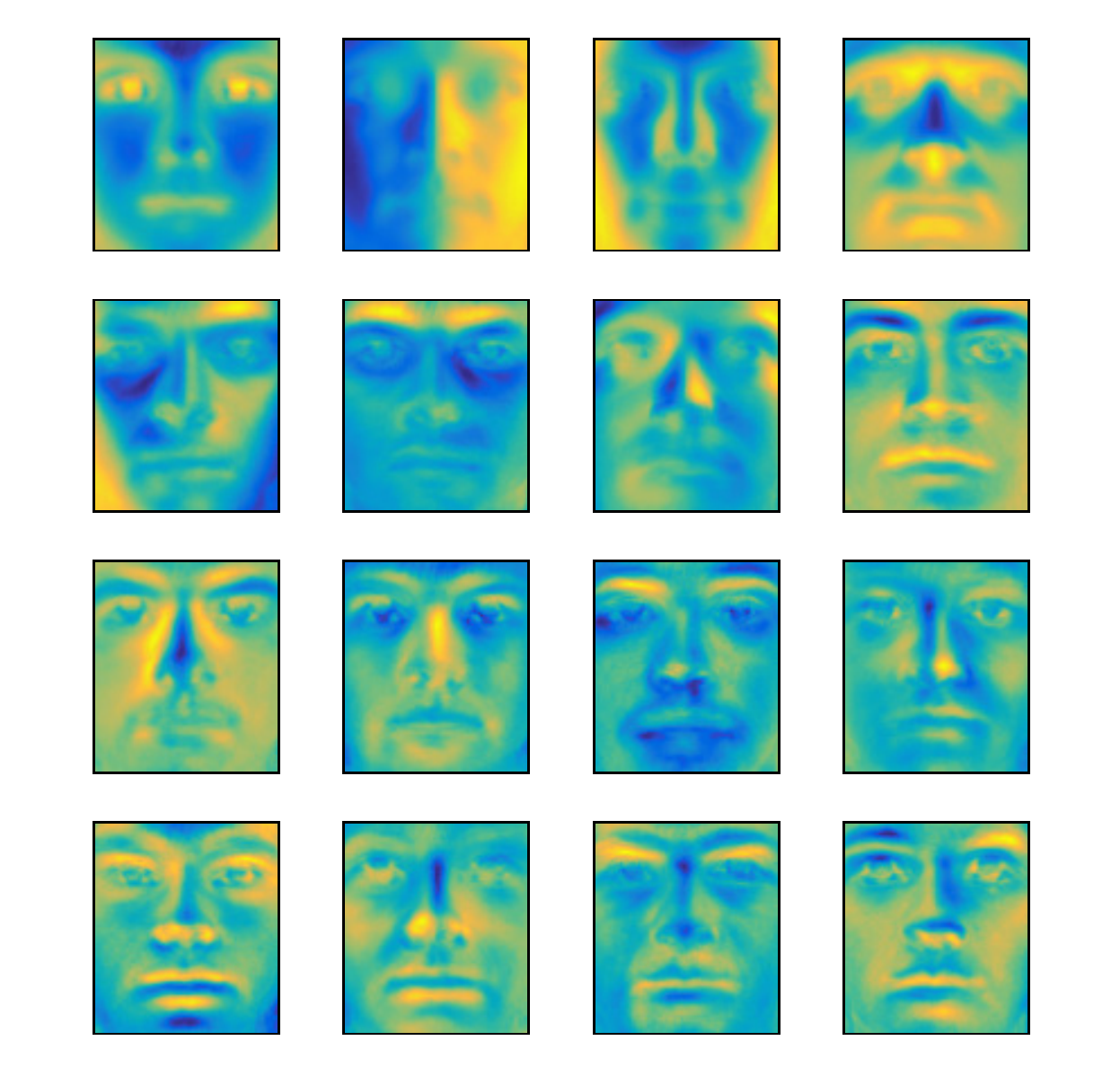}
		\caption{SVD.}
	\end{subfigure}	
	
	\caption{Dominant basis images encoding the parts-based facial features. The proposed randomized NMF algorithm faithfully captures the dominant facial features. Unlike NMF, the basis vectors computed via the SVD yield a holistic representation of the input data.}
	\label{fig:eigenfaces}
\end{figure}

In the following we use the downsampled cropped Yale face database B~\citep{yaleface}. The dataset comprises $2,410$ grayscale images, cropped and aligned. Each image is of dimension $192\times 168$. Thus, we yield a data matrix of dimension $32,256\times 2,410$ after vectorizing and stacking the images. Figure~\ref{fig:eigenfaces} shows the $16$ dominant features extracted via the deterministic and randomized HALS, which characterize some facial features. For comparison, we show also the holistic basis functions computed via the SVD. Clearly, the features extracting using the NMF are more parsimonious and better to interpret. 

The randomized algorithm achieves a substantial speed-up of a factor of about $6$, while the reconstruction error remains near-optimal, see Table~\ref{Tab:eigenfaces}. In comparison, the compressed MU algorithm performs slightly poorer overall. The compressed MU algorithm is extremely fast and the computational costs per iteration are lower than the costs of the randomized HALS algorithm. However, the MU algorithm requires a large number iterations to converge. Hence, the randomized HALS algorithms has a more favorable trade-off between speed and accuracy.
\begin{table}[!t]
	\centering
	\caption{Summary of the computational results for the Yale face database B. Baseline for speedup is deterministic HALS. The target rank is $k=16$. Note that we stopped the HALS algorithm after 500 iterations to better compare the algorithms.}
	\label{Tab:eigenfaces}	
	\scalebox{1}{
		\begin{tabular}{ l c c c c} 
			\hline 			\hline
			& \multicolumn{1}{c}{\bf Time (s)}
			& \multicolumn{1}{c}{\bf Speedup}
			& \multicolumn{1}{c}{\bf Iterations}
			& \multicolumn{1}{c}{\bf Error}									
			\\
			\cmidrule(r){1-5}
			
			\multirow{1}{*}{\rotatebox[origin=c]{0}{ \parbox{5.8cm}{Deterministic HALS}  }} 
			&  54.26	&  -  & 500 &  0.239 \\ 
			
			\multirow{1}{*}{\rotatebox[origin=c]{0}{ \parbox{5.2cm}{Randomized HALS} }} 
			&  8.93	&  6 &  500 & 0.239   \\ 
			
			\multirow{1}{*}{\rotatebox[origin=c]{0}{ \parbox{5.2cm}{Compressed MU}   }} 
			&  13.26	&  4   &  900 &  0.242   \\
			\hline \hline
		\end{tabular}
	}
\end{table}

\begin{figure}[!b]
	\centering	
	\DeclareGraphicsExtensions{.pdf}
	\includegraphics[width=0.95\textwidth]{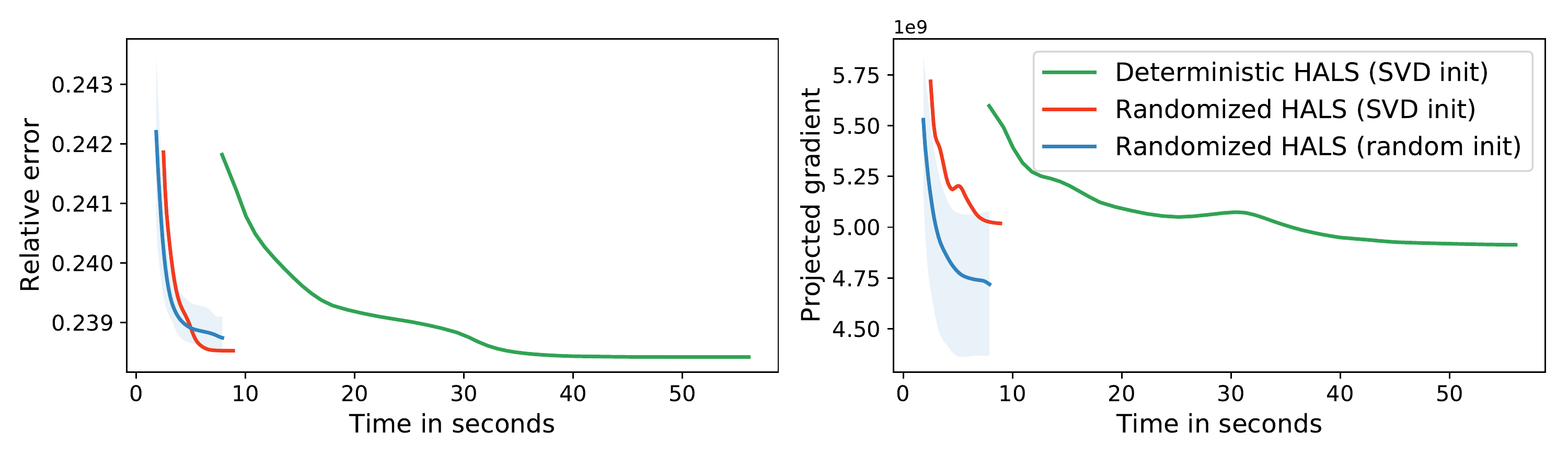}
	\caption{Relative error and projected gradient vs the computational time.}
	\label{fig:eigenfaces_time_error}
\end{figure}

\begin{figure}[!b]
	\centering	
	\vspace*{-0.2cm}
	\DeclareGraphicsExtensions{.pdf}
	\includegraphics[width=0.95\textwidth]{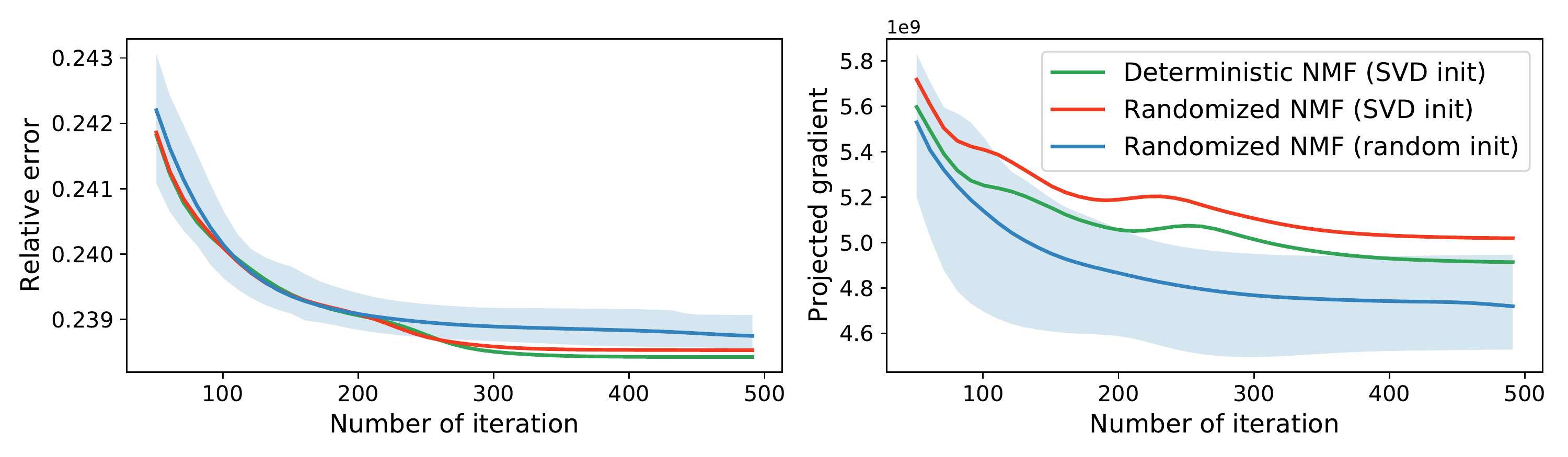}
	\caption{Relative error and projected gradient vs the number of iteration.}
	\label{fig:eigenfaces_iter_error}
\end{figure}

To better contextualize the behavior of the randomized HALS algorithm we plot the relative error and the projected gradient against the computational time in Figure~\ref{fig:eigenfaces_time_error}. The cost per iteration compared to the deterministic algorithms is substantially lower, i.e., the objective function decreases in less time.
In addition, Figure~\ref{fig:eigenfaces_iter_error} shows the results plotted against the number of iterations. Note that using the SVD for initialization increases the accuracy. While the SVD adds additional costs to the overall computational time, it is often the favorable initialization scheme~\citep{boutsidis2008svd}. 

\newpage
\subsection{Hyperspectral Unmixing}

Hyperspectral imaging is often used in order to determine what materials and underlying processes are present in a scene. It uses data collected from across the electromagnetic spectrum. 
The difficulty, however, is that pixels of hyperspectral images commonly consist of a mixture of several materials.
Thus, the aim of hyperspectral unmixing (HU) is to separate the pixel spectra into a collection of the so-called endmembers and abundances. Specifically, the endmembers represent the pure materials, while the abundances reflect the fraction of each endmember that is present in the pixel~\citep{bioucas2012hyperspectral}. The NMF represents a simple linear mixing model for blind HU in form of
\begin{equation}
p_i =  \mathbf{W} \mathbf{H}_{(:,i)} 
\end{equation}
where the $i$th pixel is denoted as $p_i$. The basis matrix $\mathbf{W}$ represents the spectral signatures of the endmembers, while the $i$th column of the weight matrix $\mathbf{H}$ represents the abundances of the corresponding pixel. 

\begin{figure}[!h]
	\centering
	\begin{subfigure}[t]{0.8\textwidth}
		\centering
		\DeclareGraphicsExtensions{.pdf}
		\includegraphics[width=1\textwidth]{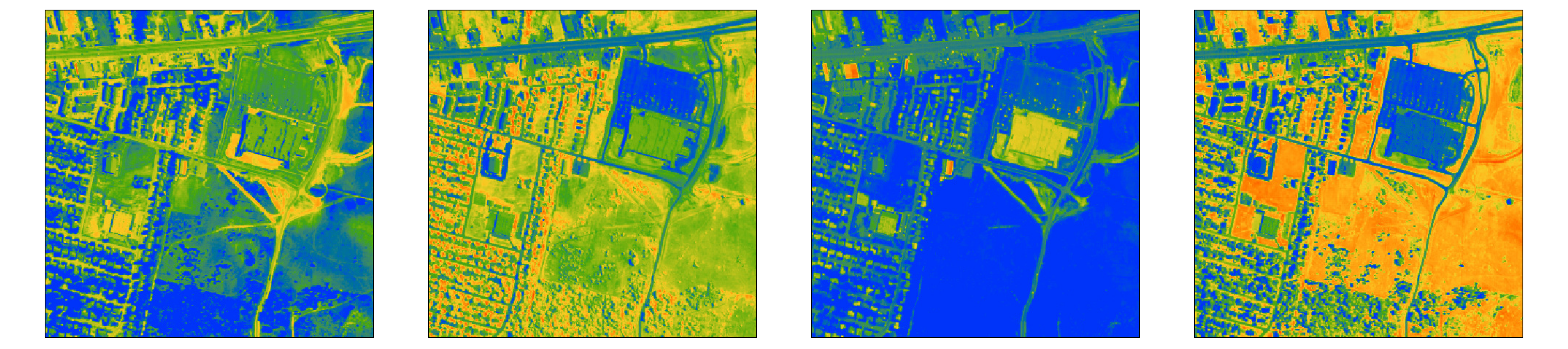}
	\end{subfigure}	
	
	\begin{subfigure}[t]{0.8\textwidth}
		\centering
		\DeclareGraphicsExtensions{.pdf}
		\vspace{+.05cm}
		\begin{overpic}[width=1\textwidth]{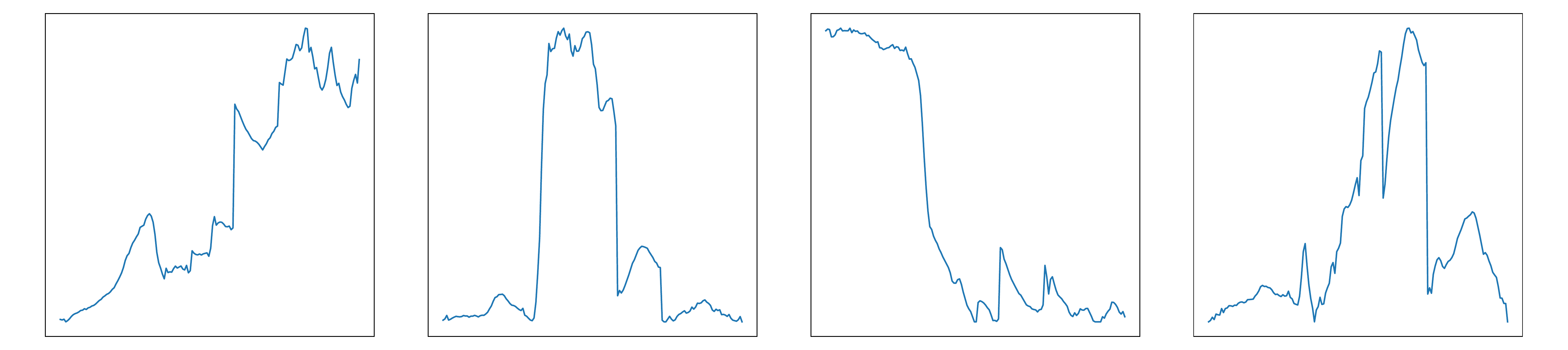}
			\put(10,23.5){\rotatebox{0}{\scriptsize asphalt}}
			\put(33,23.5){\rotatebox{0}{\scriptsize trees}}
			\put(60,23.5){\rotatebox{0}{\scriptsize roofs}}
			\put(85,23.5){\rotatebox{0}{\scriptsize grass}}
		\end{overpic}
		\caption{Deterministic HALS. }
	\end{subfigure}
	
	\vspace{+.050in}
	\begin{subfigure}[t]{0.8\textwidth}
		\centering
		\DeclareGraphicsExtensions{.pdf}
		\includegraphics[width=1\textwidth]{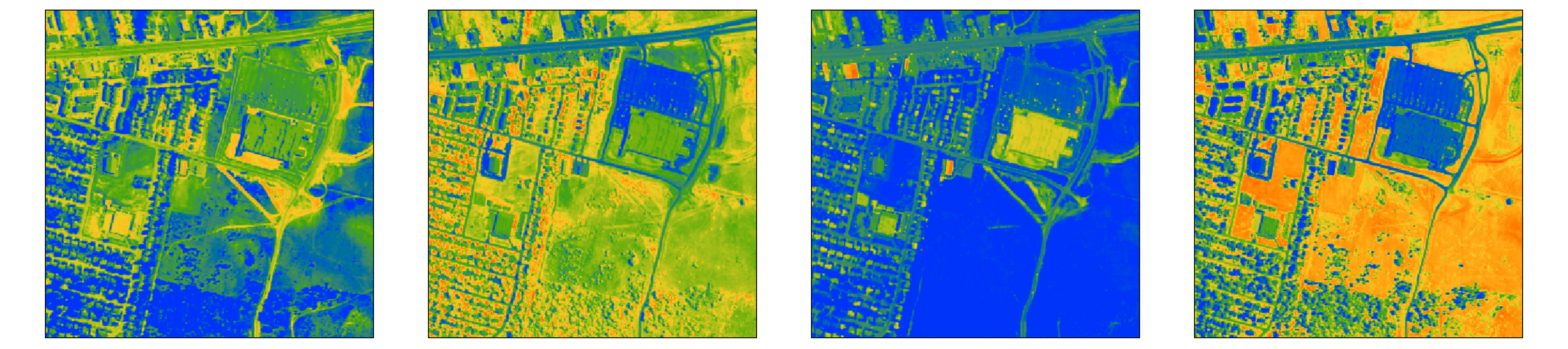}
	\end{subfigure}	
	
	\begin{subfigure}[t]{0.8\textwidth}
		\centering
		\DeclareGraphicsExtensions{.pdf}
		\vspace{+.05cm}
		\begin{overpic}[width=1\textwidth]{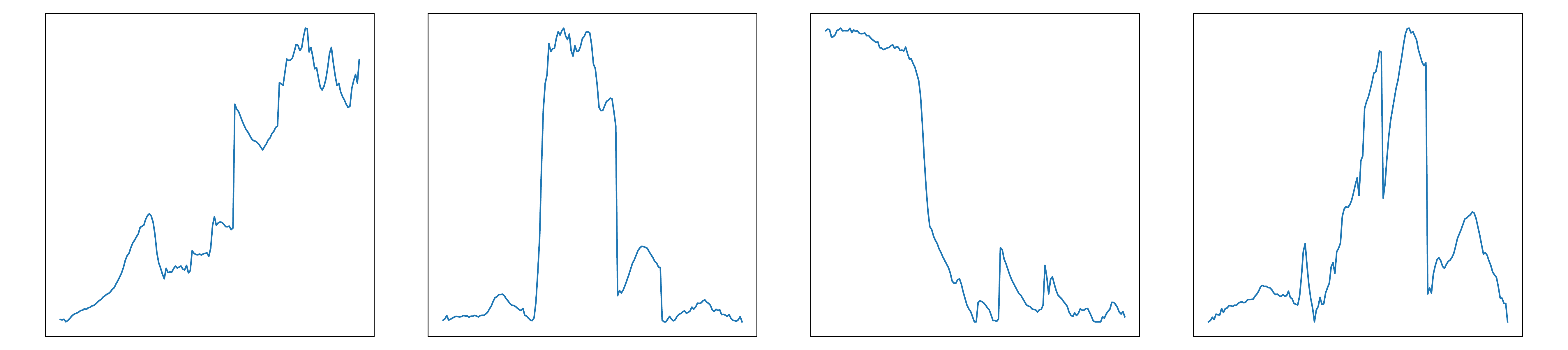}
			\put(10,23.5){\rotatebox{0}{\scriptsize asphalt}}
			\put(33,23.5){\rotatebox{0}{\scriptsize trees}}
			\put(60,23.5){\rotatebox{0}{\scriptsize roofs}}
			\put(85,23.5){\rotatebox{0}{\scriptsize grass}}
		\end{overpic}
		\caption{Randomized HALS.}
	\end{subfigure}
	
	\vspace{+.050in}	
	\begin{subfigure}[t]{0.8\textwidth}
		\centering
		\DeclareGraphicsExtensions{.pdf}
		\includegraphics[width=1\textwidth]{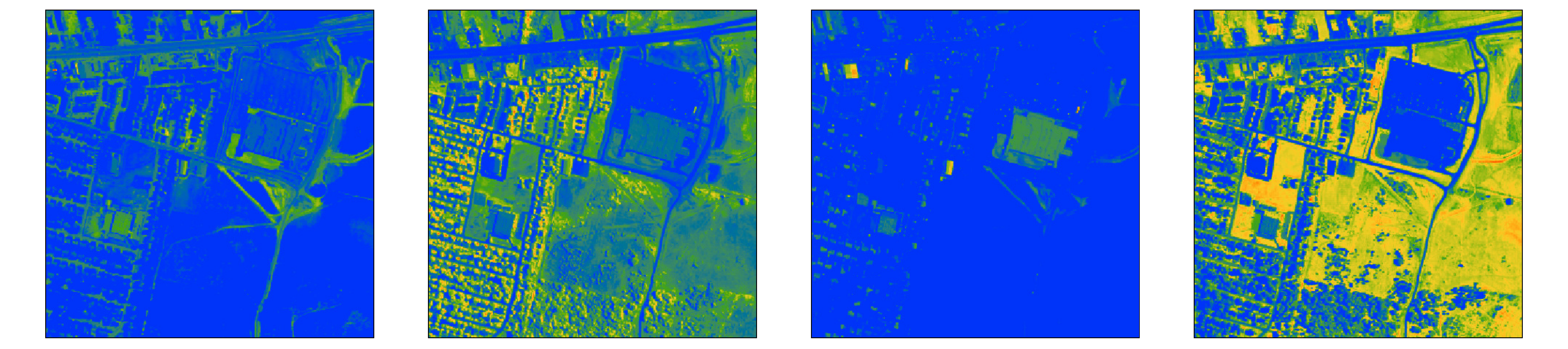}
	\end{subfigure}	
	
	\begin{subfigure}[t]{0.8\textwidth}
		\centering
		\DeclareGraphicsExtensions{.pdf}
		\vspace{+.05cm}
		\begin{overpic}[width=1\textwidth]{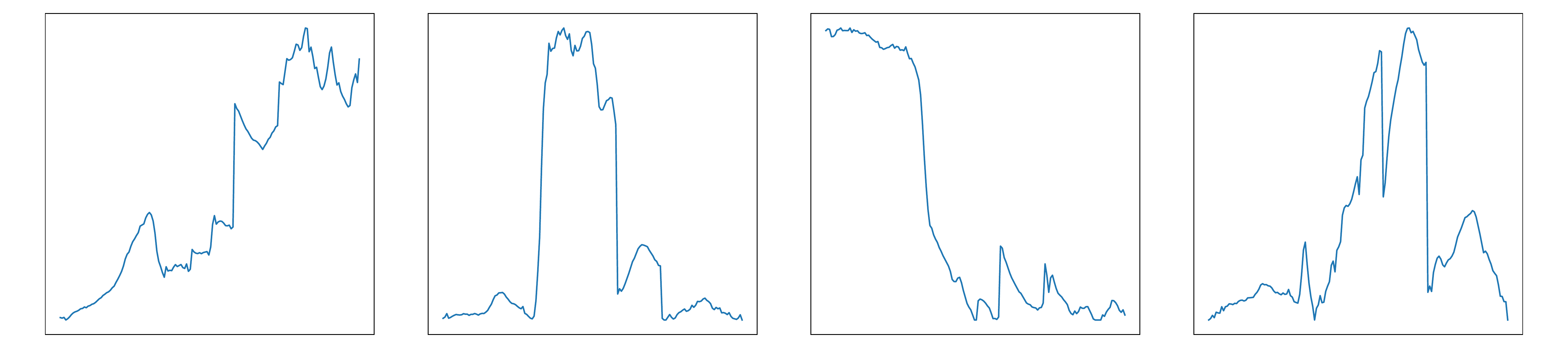}
			\put(10,23.5){\rotatebox{0}{\scriptsize asphalt}}
			\put(33,23.5){\rotatebox{0}{\scriptsize trees}}
			\put(60,23.5){\rotatebox{0}{\scriptsize roofs}}
			\put(85,23.5){\rotatebox{0}{\scriptsize grass}}
		\end{overpic}
		\caption{Regularized randomized HALS using LASSO ($\ell_1$). }
		\label{fig:hyper_sparse}
	\end{subfigure}

	\caption{Dominant basis images (endmembers) and abundance maps extracted from the `urban' hyperspectral image. (c) shows the sparse basis images using $\ell_1$ regularization. }
	\label{fig:hyper}
\end{figure}
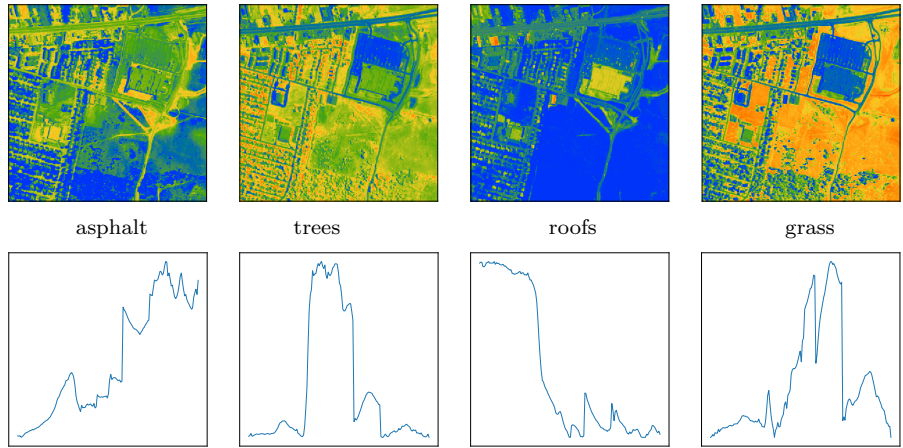
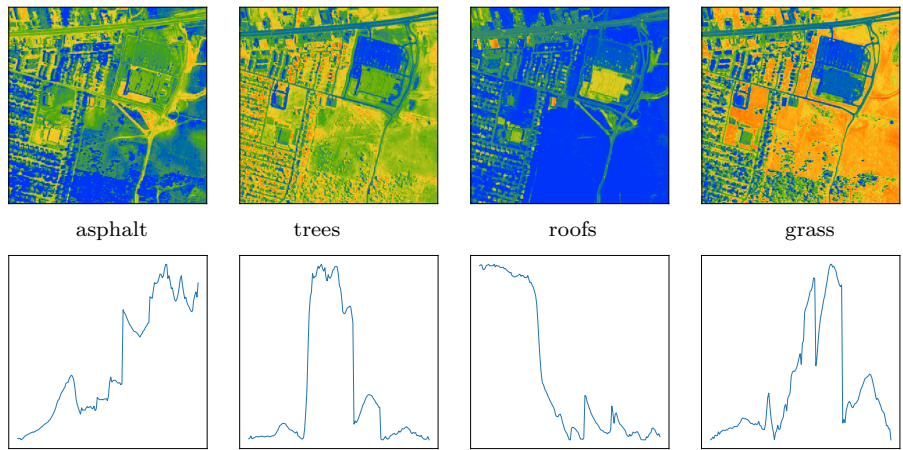
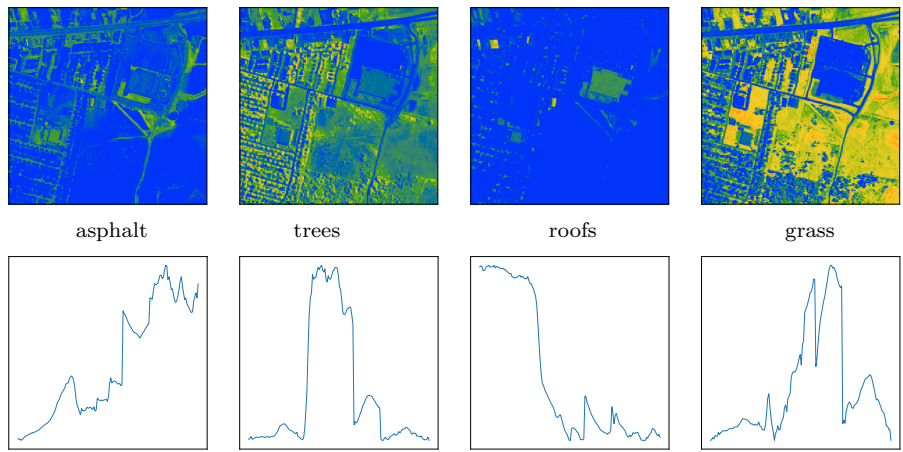

In the following we use the popular `urban' hyperspectral image for demonstration (the data are obtained from \url{http://www.agc.army.mil/}). The hyperspectral image is of dimension $307\times 307$ pixels, each corresponding to an area of $2\times 2$ meters. Further, the image consists of $210$ spectral bands; however, we omit several channels due to dense water vapor and atmospheric effects. We use only $162$ bands for the following analysis, which aims to automatically extract the four endmembers: asphalt, grass, tree and roof. 
Figure~\ref{fig:hyper} shows the $k=4$ dominant basis images reflecting the four endmembers as well as the corresponding abundance map. Both the deterministic and randomized algorithms faithfully extracted the spectral signatures and the abundance maps.
Note that we are using the SVD for initialization here, which shows a favorable performance compared to randomly initialized factor matrices. 

Table~\ref{Tab:hyper} quantifies the results. The randomized HALS and compressed MU algorithms require more iterations to converge compared to the  deterministic HALS. The speedup of randomized HALS is considerable by a factor of about $3$, whereas the MU algorithm took longer. This is, because the MU algorithm requires a larger number of iterations to converge. 

Overall, NMF is able to extract the four different endmembers, yet the modes seem to be somewhat mixed. Now, we are interested in improving the interpretability of this standard model. Therefore, we introduce some additional sparsity constraints in order to obtain a more parsimonious model. More concretely, we employ $\ell_1$ regularization to yield a sparser factor matrix $\mathbf{W}$. We control the level of sparsity via the tuning parameter $\beta$ which we set to $0.9$. Figure~\ref{fig:hyper_sparse} shows the resulting modes. Clearly, the different endmembers are less mixed and appear to be better interpretable. Note that the corresponding spectra remains the same.   

\begin{table}[!h]
	\centering
	\caption{Summary of the computational results for the hyperspectral image. Randomized HALS achieves a speedup of about $3$, while attaining the same relative error as the deterministic algorithm. The target rank is $k=4$. Baseline for speedup is deterministic HALS. }
	\label{Tab:hyper}	
	\scalebox{1}{
		\begin{tabular}{ l c c c c} 
			\hline 			\hline
			& \multicolumn{1}{c}{\bf Time (s)}
			& \multicolumn{1}{c}{\bf Speedup}
			& \multicolumn{1}{c}{\bf Iterations}
			& \multicolumn{1}{c}{\bf Error}									
			\\
			\cmidrule(r){1-5}
			
			\multirow{1}{*}{\rotatebox[origin=c]{0}{ \parbox{5.2cm}{Deterministic HALS}  }} 
			&  21.77	&  -  &  1240  &  0.0396 \\ 
			
			\multirow{1}{*}{\rotatebox[origin=c]{0}{ \parbox{5.2cm}{Randomized HALS} }} 
			&  7.23	&  3   &  1241  & 0.0396   \\ 
			
			\multirow{1}{*}{\rotatebox[origin=c]{0}{ \parbox{5.2cm}{Compressed MU}   }} 
			& 22.56 	&  -   & 2556   & 0.0398  \\
			\hline \hline
		\end{tabular}
	}
\end{table}

\begin{figure}[!t]
	\centering	
	\DeclareGraphicsExtensions{.pdf}
	\includegraphics[width=1\textwidth]{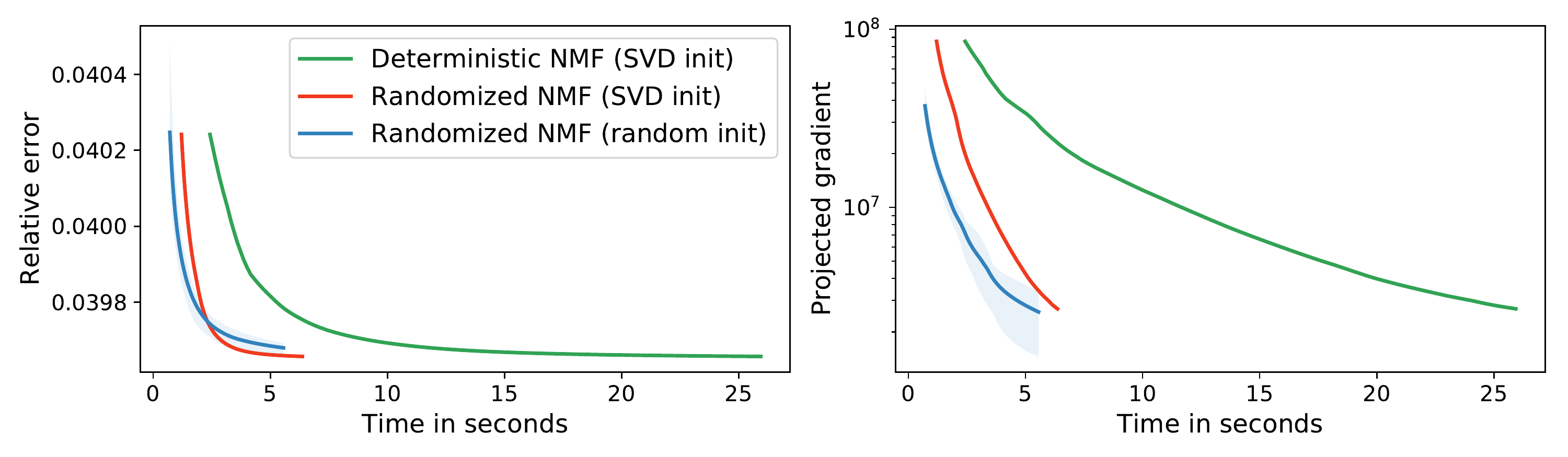}
	\caption{Relative error and projected gradient plotted vs computational time.}
	\label{fig:hyper_time_error}
\end{figure}

Next, we contextualize the performance of the randomized HALS algorithms by plotting the relative error and projected gradient vs computational time and the number of iterations, Figure~\ref{fig:hyper_time_error} and~\ref{fig:eigenfaces_iter_error}. Again, we see that the randomized algorithm faithfully converges in a fraction of the computational time required by the deterministic algorithm. Using the SVD, rather than random initialization, leads to a lower relative error on average.

\begin{figure}[!t]
	\centering	
	\DeclareGraphicsExtensions{.pdf}
	\includegraphics[width=1\textwidth]{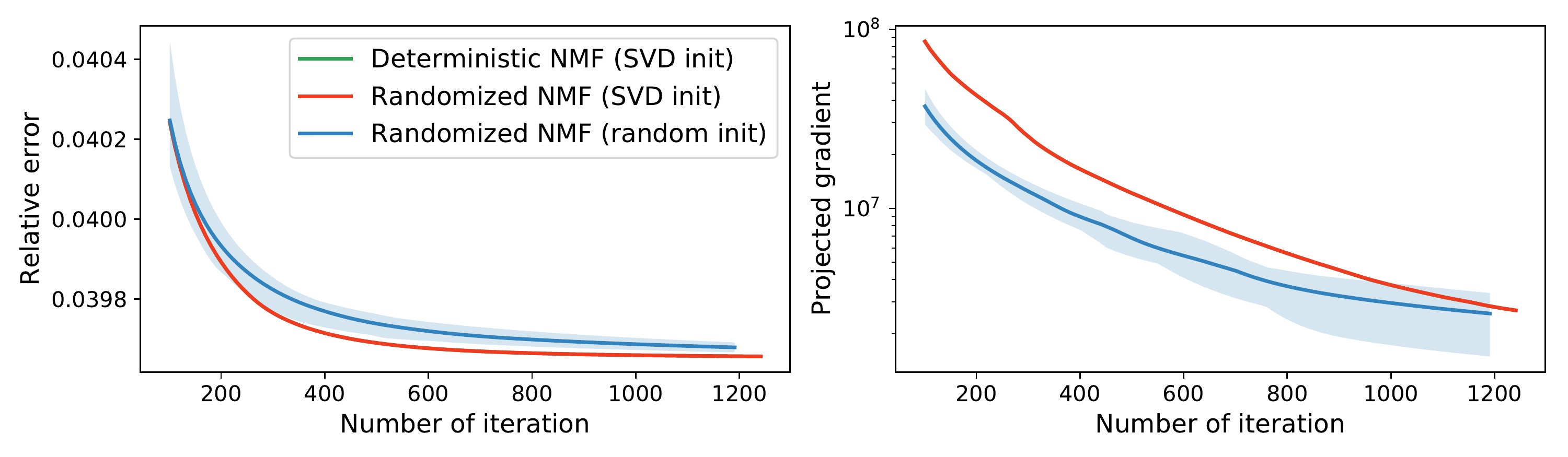}

	\caption{Relative error and projected gradient plotted vs the number of iteration.}
	\label{fig:hyper_iter_error}
\end{figure}

\subsection{Handwritten Digits}
 
In practice, it is often of great interest to extract features from data in order to characterize the underlying structure. The extracted features can then be used, for instance, for visualization or classification. Here, we are interested in investigating whether the features extracted via the randomized NMF algorithms yield a good classification performance.
 
\begin{figure}[!t]
	\centering	
	\begin{subfigure}[t]{0.24\textwidth}
		\centering
		\DeclareGraphicsExtensions{.pdf}
		\includegraphics[width=1\textwidth]{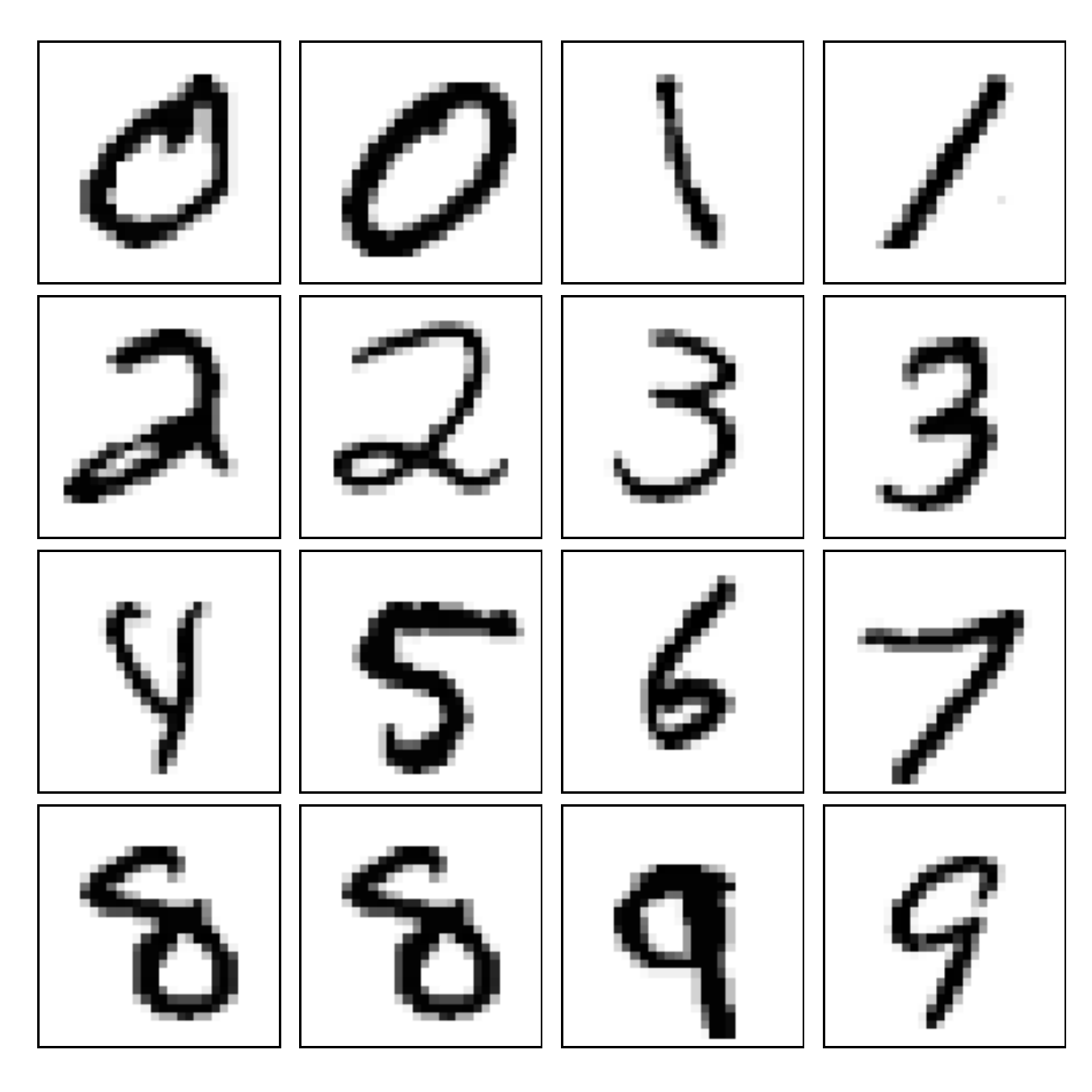}
		\caption{Sample digits. }
	\end{subfigure}	
	\begin{subfigure}[t]{0.24\textwidth}
		\centering
		\DeclareGraphicsExtensions{.pdf}
		\includegraphics[width=1\textwidth]{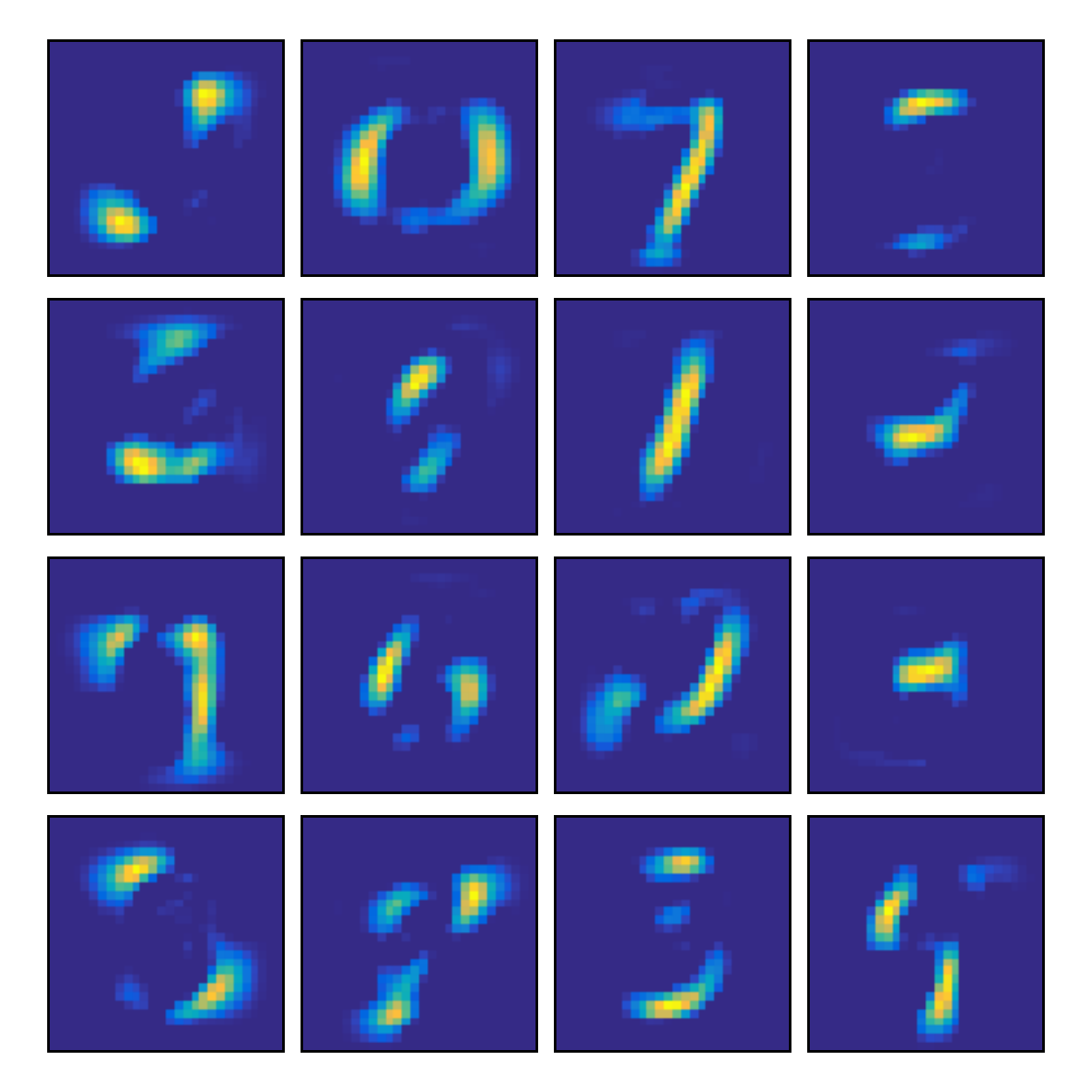}
		\caption{Deterministic NMF. }
	\end{subfigure}	
	\begin{subfigure}[t]{0.24\textwidth}
		\centering
		\DeclareGraphicsExtensions{.pdf}
		\includegraphics[width=1\textwidth]{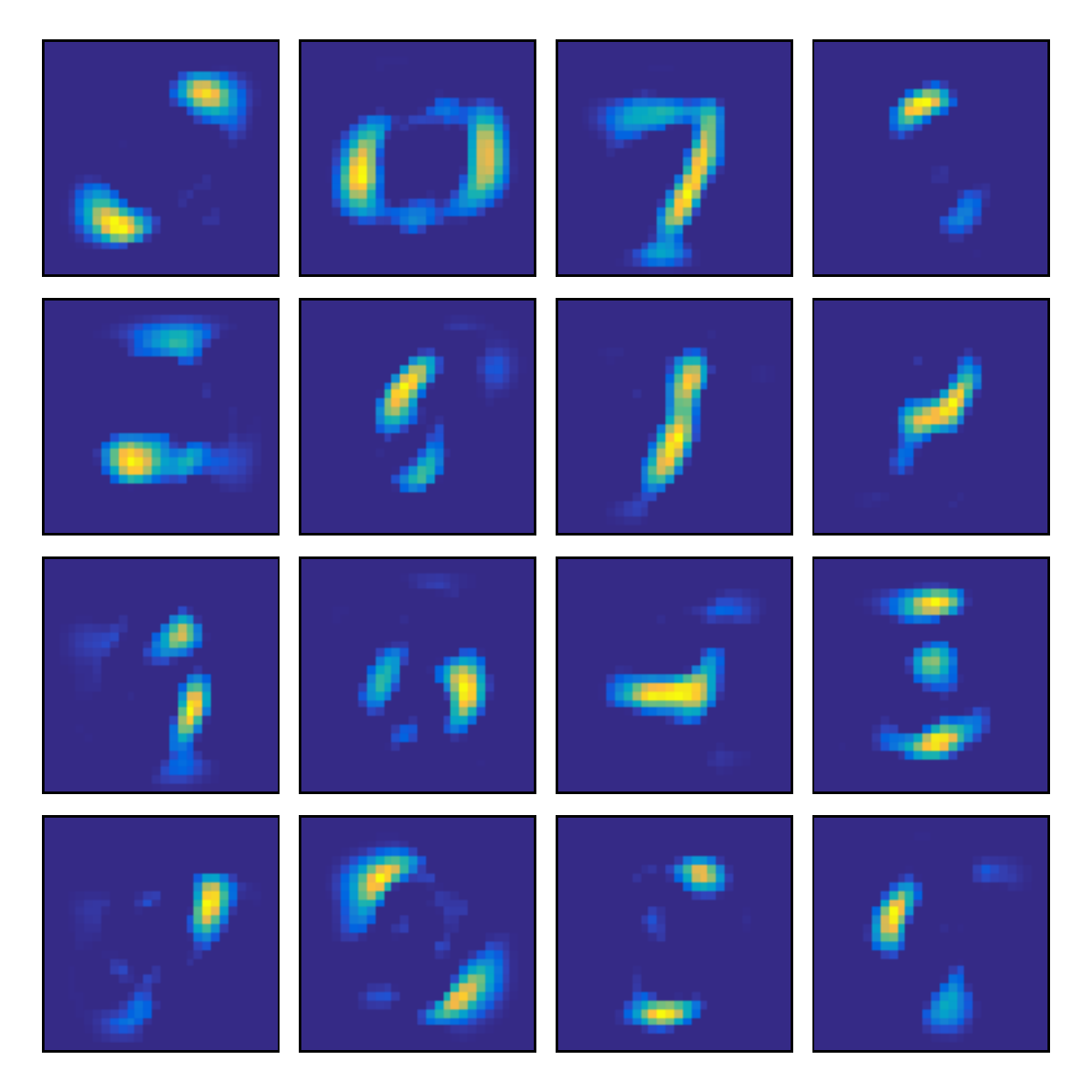}
		\caption{Randomized NMF. }
	\end{subfigure}	
	\begin{subfigure}[t]{0.24\textwidth}
		\centering
		\DeclareGraphicsExtensions{.pdf}
		\includegraphics[width=1\textwidth]{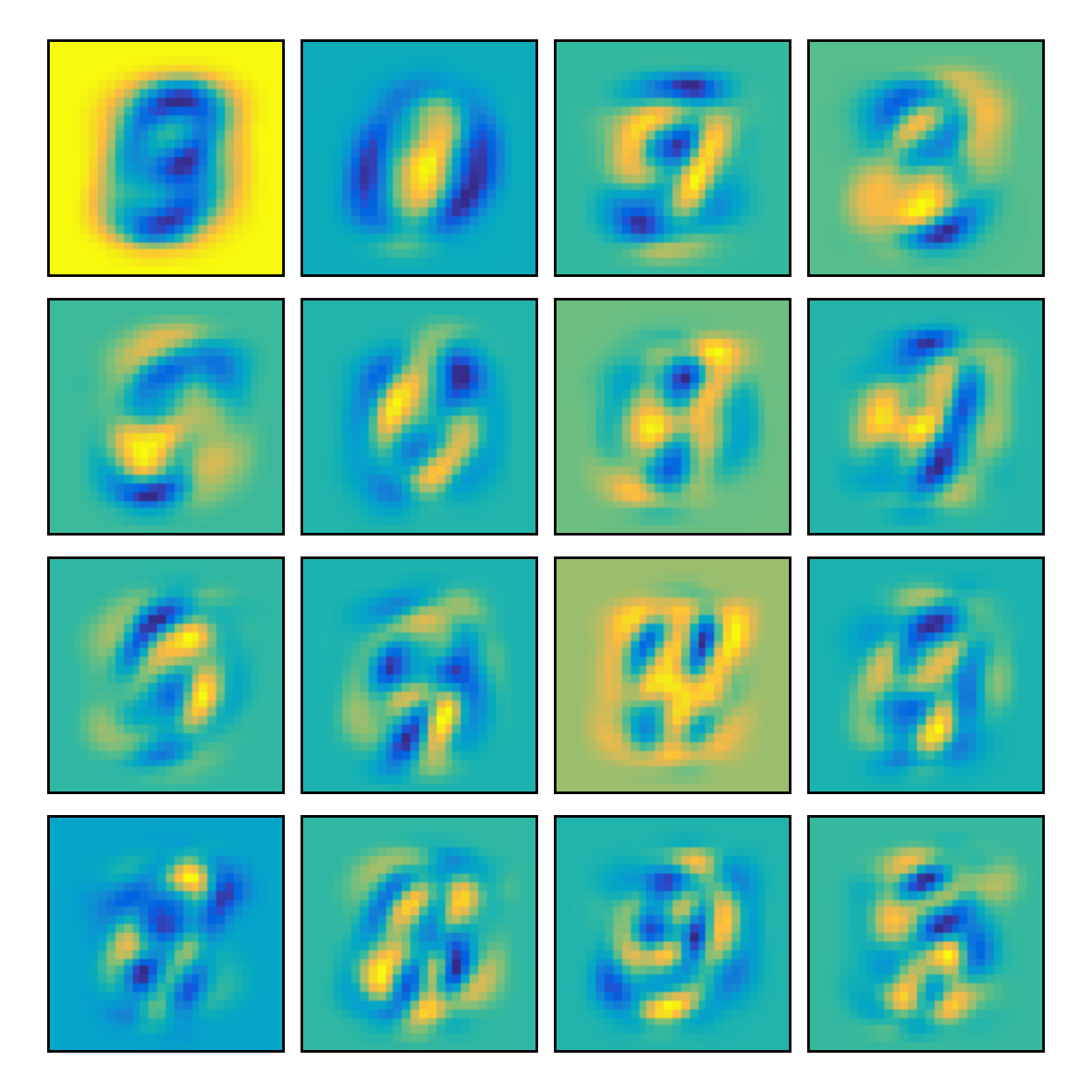}
		\caption{SVD. }
	\end{subfigure}	
	\caption{Dominant basis images extracted from the MNIST dataset (a) using the deterministic (b) and randomized HALS (c) algorithm as well as the SVD (d). Unlike the SVD, the NMF provides parts-based (sparse) features characterizing the underlying structure of the data. Note that the data are not mean centered. }
	\label{fig:mnist}
\end{figure}

In the following we use the MNIST (Modified National Institute of Standards and Technology) database of handwritten digits, which comprises $60,000$ training and $10,000$ testing images, for demonstration (the data are obtained from \url{http://yann.lecun.com/exdb/mnist/}). Each image patch is of dimension $28\times 28$ and depicts a digit between $0$ and $9$. 
Figure~\ref{fig:mnist} shows the first $16$ dominant basis images computed via the deterministic and randomized HALS as well as the SVD.
Compared to the holistic features found by the SVD, NMF computes a parts-based representation of the digits. Hence, each digit can be formed as an additive combination of the individual parts. 
Furthermore, the basis images are simple to interpret.  

Table~\ref{Tab:mnist} summarizes the computational results and shows that the randomized algorithms achieve a near-optimal reconstruction error while reducing computation. Here we limit the number of iterations to $50$ to keep the computational time low. A higher number of iterations, however, does not significantly improve the classification performance. 

Now, we investigate the question of whether the quality of the features computed by the randomized algorithm is sufficient for classification. We use the basis images to first project the data into low-dimensional space, then we use the $k$-nearest-neighbors method, with $k=3$, for classification. The results for both the training and test samples are shown in Table~\ref{Tab:classfication_results}.  

Interestingly, both the randomized and deterministic features yield a similar classification performance in terms of precision, recall, and the F1-score. 
%
%

\begin{table}[!h]
	\centering
	\caption{Summary of the computational results for decomposition of the MNIST dataset. The target rank is $k=16$. Deterministic HALS is the baseline to compute speedups. }
	\label{Tab:mnist}	
	\scalebox{1}{
		\begin{tabular}{ l c c c c} 
			\hline 			\hline
			& \multicolumn{1}{c}{\bf Time (s)}
			& \multicolumn{1}{c}{\bf Speedup}
			& \multicolumn{1}{c}{\bf Iterations}
			& \multicolumn{1}{c}{\bf Error}									
			\\
			\cmidrule(r){1-5}
			
			\multirow{1}{*}{\rotatebox[origin=c]{0}{ \parbox{5.2cm}{Deterministic HALS}  }} 
			&  4.91	&  -  &  50  &  0.547 \\ 
			
			\multirow{1}{*}{\rotatebox[origin=c]{0}{ \parbox{5.2cm}{Randomized HALS} }} 
			&  2.12	&  2.3   &  50  & 0.547   \\ 
			
			%
			
			\multirow{1}{*}{\rotatebox[origin=c]{0}{ \parbox{5.2cm}{Deterministic SVD}   }} 
			&  3.96	&  1.2   &  -  &  0.494 \\
			\hline \hline
		\end{tabular}
	}
\end{table}
\begin{table}[!h]
	\caption{Classification results of the MNIST dataset using the $k$-nearest-neighbors method, with three neighbors. The results show that the overall predictive accuracy of both the randomized and deterministic features is similar, and both are surpassed by the SVD. }
	\label{Tab:classfication_results}		
	\centering
	\scalebox{0.9}{
	\begin{subtable}{.45\textwidth}
		\centering
		\begin{tabular}{ l c c c} 
			\hline 			\hline
			& \multicolumn{1}{c}{\bf Precision}
			& \multicolumn{1}{c}{\bf Recall}
			& \multicolumn{1}{c}{\bf F1-score }
			\\
			\cmidrule(r){1-4}
			
			\multirow{1}{*}{\rotatebox[origin=c]{0}{ \parbox{3.5cm}{Deterministic HALS}  }} 
			&  0.97   &   0.97   &   0.97   \\ 
			
			\multirow{1}{*}{\rotatebox[origin=c]{0}{ \parbox{3.2cm}{Randomized HALS} }} 
			&  0.97   &   0.97   &   0.97   \\ 
			
			%
			
			\multirow{1}{*}{\rotatebox[origin=c]{0}{ \parbox{3.2cm}{Deterministic SVD}   }} 
			&  0.98   &   0.98   &   0.98 \\
			\hline \hline
		\end{tabular}
	\caption{Training data.}	
	\end{subtable}}
	\hspace*{2.5cm}
	\scalebox{0.9}{
	\begin{subtable}{.45\textwidth}
		\centering
		\begin{tabular}{ l c c c} 
			\hline 			\hline
			& \multicolumn{1}{c}{\bf Precision}
			& \multicolumn{1}{c}{\bf Recall}
			& \multicolumn{1}{c}{\bf F1-score }
			\\
			\cmidrule(r){1-4}
			
			\multirow{1}{*}{\rotatebox[origin=c]{0}{   }} 
			&  0.95   &   0.95   &   0.95   \\ 
			
			\multirow{1}{*}{\rotatebox[origin=c]{0}{  }} 
			&  0.95   &   0.95   &   0.95    \\ 
			
			%
			
			\multirow{1}{*}{\rotatebox[origin=c]{0}{ \  }} 
			&  0.96   &   0.96   &   0.96 \\
			\hline \hline
		\end{tabular}
	\caption{Test data.}			
	\end{subtable}}%
\end{table}

\newpage
\subsection{Computational performance}
We use synthetic nonnegative data to contextualize the computational performance of the algorithms. Specifically, we construct low-rank matrices consisting of nonnegative elements drawn from the Gaussian distribution.
 
\begin{figure}[!b]
	\centering
	\begin{subfigure}[t]{0.49\textwidth}
		\centering
		\DeclareGraphicsExtensions{.pdf}
		\includegraphics[width=1\textwidth]{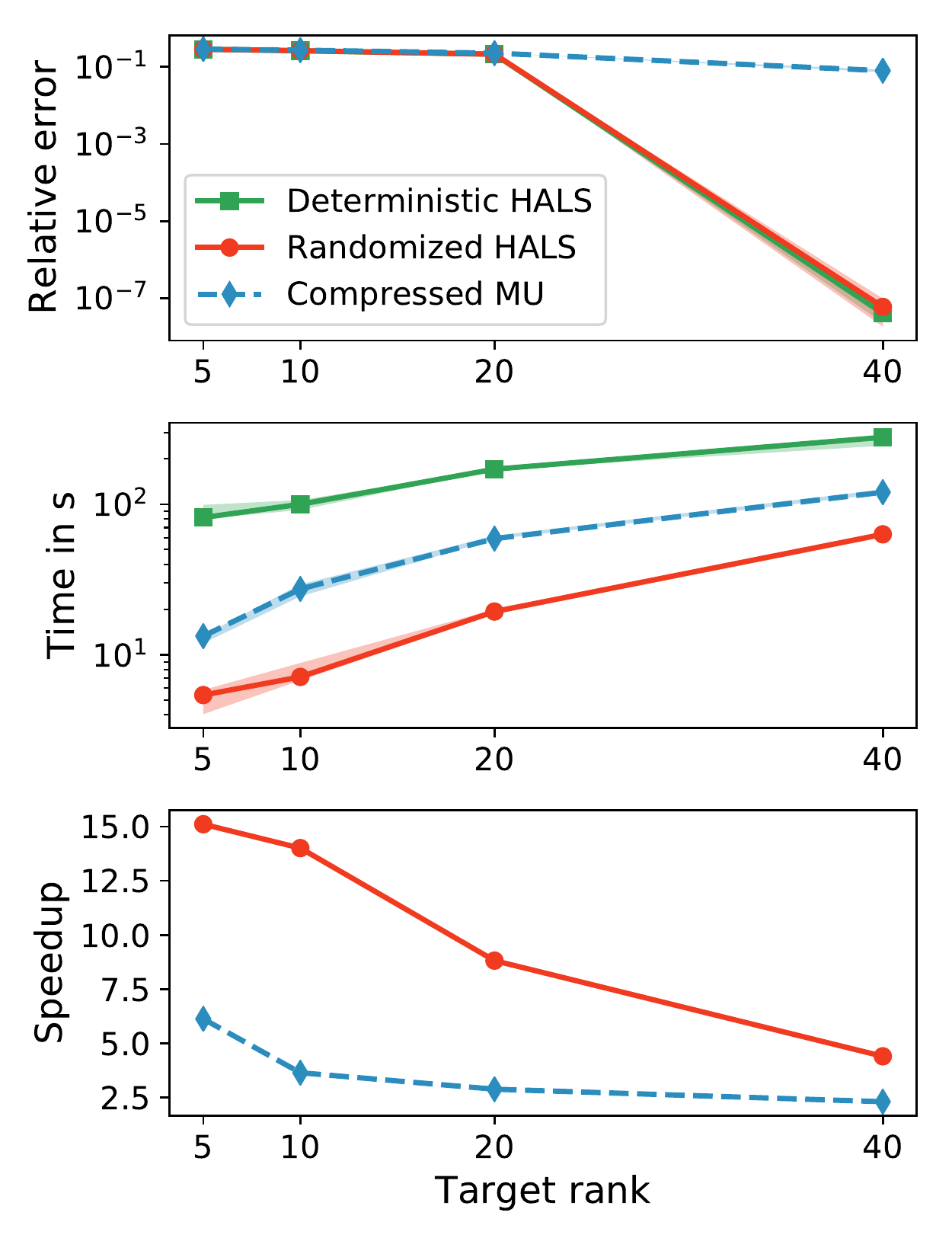}
		\caption{Dimension: $100,000\times 5,000$.}
	\end{subfigure}	
	\hspace{-.15in}
	\begin{subfigure}[t]{0.49\textwidth}
		\centering
		\DeclareGraphicsExtensions{.pdf}
		\includegraphics[width=1\textwidth]{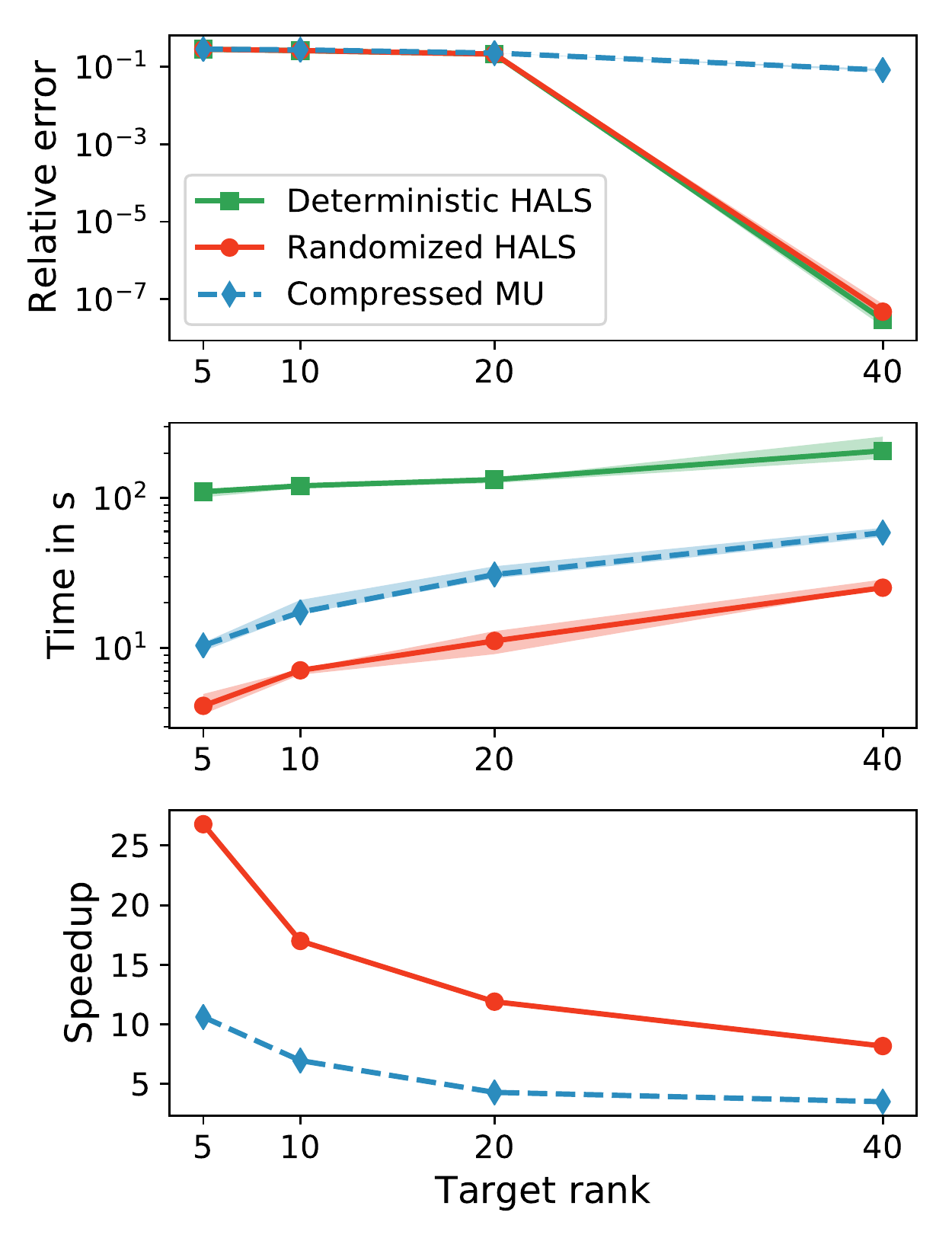}
		\caption{Dimension: $25,000\times 25,000$.}
	\end{subfigure}	
	\caption{The randomized HALS algorithm outperforms on both tall-and-skinny (a), and fat (b) synthetic data matrices of rank $50$. The compressed MU algorithm fails to converge in (b). Baseline for speedup is deterministic HALS. }
	\label{fig:performance}
\end{figure}

First, we compute the NMF for low-rank matrices of dimension $100,000\times 5,000$ and $25,000\times 25,000$, both of rank $r=40$.
Figure~\ref{fig:performance} shows the relative error, the timings, and the speedup for varying target ranks $k$, averaged over 20 runs. First, we notice that the randomized HALS algorithm shows a significant speedup over the deterministic HALS algorithm by a factor of $3$ to $25$, while achieving a high accuracy. Here, we have limited the maximum number of iterations to $200$ for both the randomized and deterministic HALS algorithm. If required, an even higher accuracy could be achieved by allowing for a larger number of iterations. 

The MU algorithm is known to require a larger number of iterations. 
Thus, we have allowed for a maximum number of iterations of $1,000$. 
Despite the large number of iterations, the compressed MU algorithms shows a patchy performance in comparison.
While the results for small target ranks are satisfactory, the algorithm has difficulties in approximating factors of larger ranks accurately.
In both cases, the compressed MU algorithm does not converge.
%

%

%
%

%
\begin{figure}[!t]
	\centering	
	\DeclareGraphicsExtensions{.pdf}
	\includegraphics[width=0.99\textwidth]{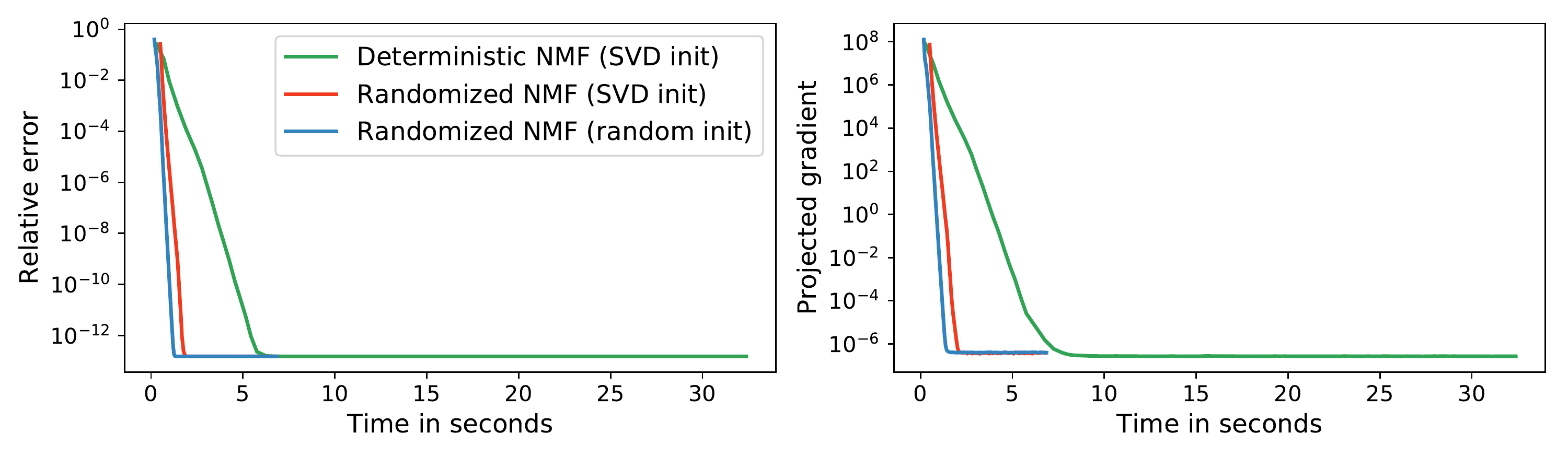}
	\caption{Relative error and projected gradient plotted vs computational time.}
	\label{fig:rand_time_error}
\end{figure}

\begin{figure}[!t]
	\centering	
	\DeclareGraphicsExtensions{.pdf}
	\includegraphics[width=0.99\textwidth]{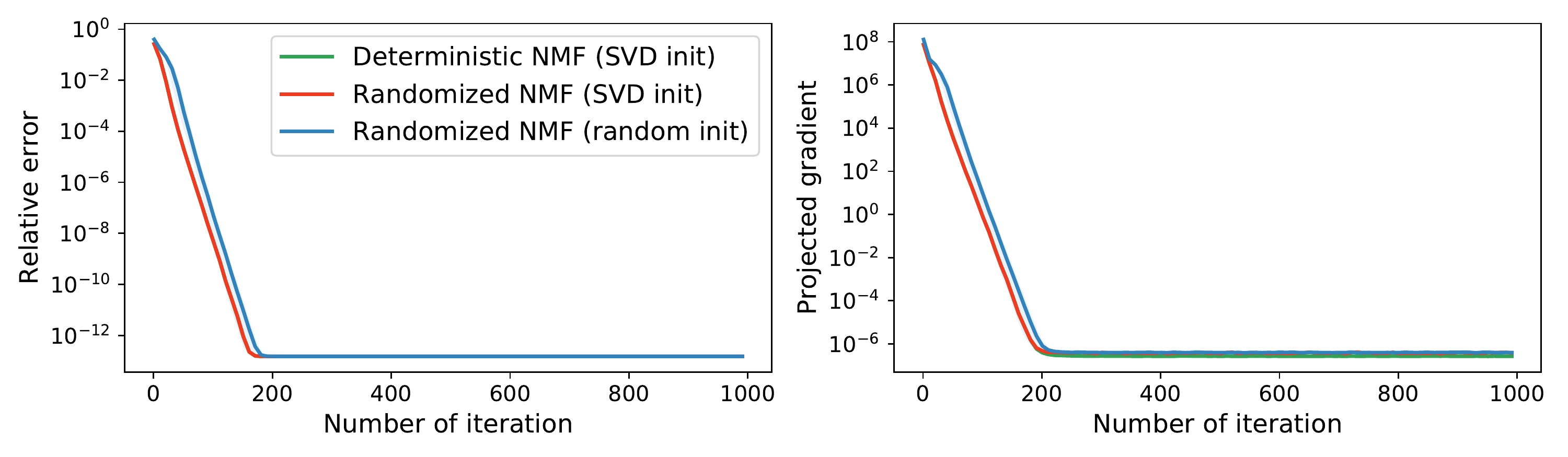}

	\caption{Relative error and projected gradient plotted vs the number of iteration.}
	\label{fig:rand_iter_error}
\end{figure}

Figure~\ref{fig:rand_time_error} and~\ref{fig:rand_iter_error} shows the convergence behavior for a random generated low-rank matrix with dimensions $5,000\times 5,000$. Like the deterministic algorithm, the randomized HALS algorithms approximates the data with nearly machine-precision. 
While using the SVD for initialization requires some additional computational costs, the accuracy is slightly better.
\section[Conclusion]{Conclusion}\label{sec:conclusion}

Massive nonnegative data poses a computational challenge for deterministic NMF algorithms. However, randomized algorithms are a promising alternative for computing the approximate nonnegative factorization. 
The computational complexity of the proposed randomized algorithm scales with the target rank rather than ambient dimension of the measurement space. Hence, the computational advantage becomes pronounced with increasing dimensions of the input matrix.

The randomized HALS algorithm substantially reduces the computational demands while achieving near-optimal reconstruction results. 
Thereby, the trade-off between precision and speed can be controlled via oversampling and utilizing power iterations. 
We prose $p=20$ and the computation of $q=2$ subspace iterations as default values. 
These settings show a good performance throughout all our experiments. Overall, the randomized HALS algorithm shows considerable advantages over the deterministic HALS and the previously proposed compressed MU algorithm. In addition, regularized HALS can help to compute more interpretable modes.

Future research will investigate a GPU-accelerated implementation of the proposed randomized HALS algorithm. Furthermore, the presented ideas can be applied to nonnegative tensor factorization using the randomized framework proposed by~\cite{Erichson2016rcpd}.

\newpage
\section*{Acknowledgments}
We would like to express our gratitude to the two anonymous reviewers for their helpful feedback which allowed us improve the manuscript.
NBE and JNK acknowledge support from an SBIR grant through SURVICE Inc. (FA9550-17-C-0007)


\appendix
\section{QB Decomposition}\label{appendix:qb}

If the data matrix $\mathbf{A}$ fits into fast memory, the QB decomposition can be computed efficiently using BLAS-3 operations. However, in a big data environment we face the challenge that the data matrix is to big to fit into fast memory. The randomized scheme allows to build the smaller matrix $\mathbf{B}$ by simply iterating over the columns or blocks of columns of $\mathbf{A}$, once at a time. For instance, the HDF5 file system allows a handy framework to access subsets of columns of a data matrix. 

Algorithm~\ref{alg:rqb} shows a prototype implementation of the QB decomposition. Without additional power iterations, the algorithm requires two passes over the entire data matrix. Each additional power iteration requires one additional pass. Note, that in practice it is more efficient to read in blocks, rather than just a single column. Further, the for-loops in line (4), (8) and (12) can be executed in parallel. 

\begin{algorithm}[!b]
	\scalebox{0.99}{\fbox{		
			\begin{minipage}{210mm}
				\begin{tabbing}
					\hspace{2mm} \= \hspace{5mm} \= \hspace{5mm} \= \hspace{5mm} \= \hspace{50mm} \=\kill
					\textbf{Input:} Input matrix $\mathbf{A}$ with dimensions $m\times n$, and target rank $k<\text{min}\{m,n\}$.\\[1mm]
					\textbf{Optional:} Parameters $p$ and $q$ to control oversampling, and the power scheme.\\[3mm] 
					
					\textbf{function} $\texttt{rqb}(\mathbf{A}, k, p, q)$\\[3mm]

					(1)  \> \> $l = k + p$ \> \> \> {\color{blue}$\textrm{slight oversampling}$} \\[1mm]
					
					(2)  \> \> $\mathbf{\Omega} = \texttt{rnorm}(n,l)$ \> \> \> {\color{blue}$\textrm{generate random test matrix}$}\\[1mm]
					
					(3)  \> \> $\mathbf{Y} = \texttt{zeros}(m,l)$ \> \> \> {\color{blue}$\textrm{initialize matrix}$}\\[1mm]						
					
					(4)  \> \> \textbf{for} $j = 1,\dots,n$ \> \> \> {\color{blue}$\textrm{Read-in } j\textrm{th column}$} \\[1mm]					
					
					(5)  \> \> \> $\mathbf{Y} = \mathbf{Y} + \mathbf{A}(:,j) \mathbf{\Omega}$ \> \> {\color{blue}$\textrm{update sketch}$}\\[3mm]
					
					(6)  \> \> \textbf{for} $j = 1,\dots,q$ \> \> \> {\color{blue}$\textrm{perform q iterations}$} \\[1mm]
					
					(7)  \> \> \> $\left[\mathbf{Q},\sim\right] = \texttt{qr}(\mathbf{Y})$ \> \> {\color{blue}$\textrm{economic QR}$}\\[1mm]
					
					(8)  \> \> \> \textbf{for} $j = 1,\dots,n$ \>  \> {\color{blue}$\textrm{Read-in } j\textrm{th column}$} \\[1mm]					
					
					(9)  \> \> \> \>$\mathbf{Y} = \mathbf{A}(:,j) (\mathbf{\mathbf{A}(:,j)^\top \mathbf{Q}})$ \\[3mm]

					(10)  \> \> $\left[\mathbf{Q},\sim\right] = \texttt{qr}(\mathbf{Y})$ \> \> \> {\color{blue}\textrm{form orthonormal basis}}
					\\[1mm]

					(11)  \> \> $\mathbf{B} = \texttt{zeros}(l,n)$ \> \> \> {\color{blue}$\textrm{initialize matrix}$}\\[1mm]

					(12)  \> \> \textbf{for} $j = 1,\dots,n$ \> \> \> {\color{blue}$\textrm{Read-in } j\textrm{th column}$} \\[1mm]					
					
					(13)  \> \> \> $\mathbf{B}(:,j) = \mathbf{Q}^\top \mathbf{A}(:,j)$ \> \> {\color{blue}$\textrm{project to low-dimensional space}$}\\[3mm]

					\textbf{Return:} $\mathbf{Q}\in \mathbb{R}^{m\times l}$, $\mathbf{B}\in \mathbb{R}^{l\times n}$
				\end{tabbing}
			\end{minipage}}}
			\centering
			\vspace{+.15in}
			\caption{A randomized QB decomposition algorithm.}
			\label{alg:rqb}
		\end{algorithm}

\newpage
\begin{spacing}{0.98}\small
\bibliography{bib}   
\end{spacing}

\end{document}